\documentclass{article}

\usepackage{PRIMEarxiv}

\usepackage[utf8]{inputenc} % allow utf-8 input
\usepackage[T1]{fontenc}    % use 8-bit T1 fonts
\usepackage{hyperref}       % hyperlinks
\usepackage{url}            % simple URL typesetting
\usepackage{booktabs}       % professional-quality tables
\usepackage{amsfonts}       % blackboard math symbols
\usepackage{nicefrac}       % compact symbols for 1/2, etc.
\usepackage{microtype}      % microtypography
\usepackage{lipsum}
\usepackage{fancyhdr}       % header
\usepackage{graphicx}       % graphics
\usepackage{tabularx}       % table
\usepackage{listings}       % code block
\graphicspath{{media/}}     % organize your images and other figures under media/ folder

\title{Macro-Queries: An Exploration into Guided Chart Generation from High Level Prompts
}

\author{
  Christopher J. Lee, Giorgio Tran, Roderick Tabalba, Jason Leigh \\
  \\
  Laboratory for Advanced Visualization and Applications \\
  University of Hawai`i at Manoa \\
  Honolulu\\
  \\
  \texttt{\{clee48, ttran2, tabalbar, leighj\}@hawaii.edu} \\
   \texttt{lava.hawaii.edu}
  \And
  Ryan Longman \\
  East-West Center and Water Resources Research Center \\
  University of Hawai`i at Manoa \\
  Honolulu\\
  \texttt{rlongman@hawaii.edu} \\
}

\begin{document}
\maketitle

\begin{abstract}
This paper explores the intersection of data visualization and Large Language Models (LLMs).
Driven by the need to make a broader range of data visualization types accessible for novice users, we present a guided LLM-based pipeline designed to transform  data, guided by high-level user questions (referred to as macro-queries), into a diverse set of useful visualizations. This approach leverages various prompting techniques, fine-tuning inspired by Abela's Chart Taxonomy, and integrated SQL tool usage. 
\end{abstract}

% keywords can be removed
\keywords{Large Language Model \and LLM \and Prompt Engineering \and SQL \and Macro Queries \and  Data \and Visualization \and Charts \and High Level Queries}

%understanding and generation 
\section{Introduction and Background}
With the recent surge in popularity of Large Language Models (LLMs), providing unprecedented advancements in the domain of natural language and motivated by the imperative to make information accessible to novices; we seek to explore the junction between data visualization and artificial intelligence.  This is achieved through the integration of LLM capabilities into chart production via various prompting techniques and finetuning,  thereby facilitating the transformation of high-level user prompts, which we denote as macro-queries, into actionable sequences that result in the automated production of charts.  Our systematic approach leverages the capabilities of LLMs at each stage; in an effort to decompose complex user queries and extract user intentions to formulate contextually accurate responses, aligned with user provided spreadsheet data, that culminate in the generation of insightful visual data representations.

% Among the various outlets for...

%In our research with Chatgpt 3.5, the LLM, when asked to show relationships between data without explicitly declaring chart types, only chose to generate charts from the following pool: (...).  Our approach to extract the ability to generate charts outside the limited scope involves few shot 

% In domain of data visualization, where a significant of chart variants exist, we chose to use Andrew Abela's chart taxonomy due to xyz.

% With the recent surge in popularity of Large Language Models (LLMs) such as....

% With the surge of large language models, we attempt to improve their performance 
% seeking methodologies to improve LLM performance on charts...

% inspired by quick implementations...?

% The following constrained LLM agent architect-ed aims to include both accuracy and verifiability...

% benefits

% \subsection{Macro-Queries}
% \paragraph{Macro-Queries:}
% \subsection{Motivation}
 
% MORE  ABOUT characteristics OF MACROQUEIRIES
\subsection{Macro-Queries}
We prognosticate the increasing importance of distinctively classifying macro-queries from  decades of observed interactions with domain scientists, government planners, policy makers and the general public. One particular example is within the Change Hawai`i Project \cite{changehi}, a \$20M National Science Foundation to study the impacts of climate change in Hawai`i. The project is comprised of climate science researchers,  computer scientists, and indigenous practitioners, developing the Hawai`i Climate Data Portal \cite{hcdp}\cite{longman2024hcdp}\cite{mclean2021building}\cite{mclean2020hawaii}- a database that collects all climate-related data for the State of Hawai`i.  The feedback associated among stakeholders  persistently showed heightened interest in requesting high level queries.  Examples include "when is the right time to plant crops", "which direction is the fire likely to spread", or "what is the likelihood of a flood in the next 5 years". 

Therefore we coined the term macro-queries in the context of data exploration.  Macro-queries entails a broad or high level request for knowledge about the data, typically without directly referencing data attributes primed for manipulation, and for which fulfilling the request may require a complex set of steps that may include a combination of, but not limited to data transformations (i.e., aggregation, filtering, ...) , planning, or web searches.  As an example, when referring to the well known dataset associated with cars \cite{carprice}, a macro-query could be "Which car is best for camping?" or "Show me the best-bang for buck cars."  This does not refer to any specific desired transformations or attributes from the dataset and requires logical abstractions with creativity to determine metrics for "best."  Most importantly, macro-queries should be imagined as a spectrum from non-macro-queries, where intentions are clear and no guess work can be done, to macro-queries where there are a plethora of possible approaches and solutions for resolving the request.  For further clarification see Table~\ref{table:macroquery}.

However macro-queries, which is a category of user prompt, are not to be confused with  meta-prompts\cite{suzgun2024metapromptingenhancinglanguagemodels} which focuses on crafting or designing of prompts to guide AI systems to generate more effective responses. The former focuses on the user's scientific question whereas the latter focuses on the mechanics of how to answer the question.

% In the event of a temperature data set where two attributes maybe temperature and date; asking  is not a macroquery,

% whats the most affordable vehicle?
\begin{table}
\caption{Examples of Macro-Queries vs Non-Macro-Queries; Attributes: [car name, price, city mpg, highway mpg]}
\label{table:macroquery}
\begin{tabularx}{1\textwidth} { 
  | >{\raggedright\arraybackslash}X 
  | >{\raggedright\arraybackslash}X 
  | >{\raggedright\arraybackslash}X | }
 \hline
 \begin{center} \textbf{Example} \end{center} & \begin{center}  \bf Is macro-query? \end{center} & \begin{center} \bf Reason \end{center} \\
\hline
 "show me a bar chart of car names and their prices sorted by price." & \center no & The user's request distinctly refers to the data attributes in the data set.  Furthermore, the user defines the chart and the transformations. There is no need for guess work. \\
\hline
 "show me the car names with the lowest price" & \center no & The user's request distinctly refers to the data attributes in the data set. The only ambiguities pertains to chart selection and specifying a cutoff point for "lowest" price. \\
\hline
 "show me the car name for the most affordable car" & \center yes & Affordability depends on unknown factors inclusive of user's income which must be inferred and can result in multiple answers.  However the user did specify a desire for the car name attribute.  Therefore, this is mostly a macro-query.\\
\hline
 "which is the most affordable car?" & \center yes & The query does not directly reference any data attributes and requires the data to be sorted.  Multiple interpretations can be made about affordability. (e.g., most affordable upfront costs or long term costs)\\
\hline
\end{tabularx}
\end{table}

% \begin{tabularx}{0.8\textwidth} { 
%   | >{\raggedright\arraybackslash}X 
%   | >{\raggedright\arraybackslash}X 
%   | >{\raggedright\arraybackslash}X | }
%  \hline
%  example & is macro-query? & reason \\
%  \hline
%  "what is the temperature between n and m dates" & no & The prompt directly asks for the input attribute of date.  \\
%  \hline
%  "what is going to be the temperature tomorrow?" & yes & The attribute referenced is the output attribtue and not input attribute.  Furthermore, it is requesting a predictive data-point.  \\
%  \hline
%  "what is going to be the temperature tomorrow?" & yes & The attribute referenced is the output attribtue and not input attribute.  \\
% \hline
% \end{tabularx}

% \begin{tabularx}{0.8\textwidth} { 
%   | >{\raggedright\arraybackslash}X 
%   | >{\centering\arraybackslash}X 
%   | >{\raggedleft\arraybackslash}X | }
%  \hline
%  example & is macro-query? & reason \\
%  \hline
%  item 21  & item 22  & item 23  \\
% \hline
% \end{tabularx}

% Therefore, "show me a chart of car brands vs mpg" and "Which is the best engine type?" should not be considered macro-queries due to direct references to attributes.

% We changed compontents of components to tree map as it represented the same thing, but clearer
% in conjunction with

\subsection{LLM Approach}
% \paragraph{LLM Pipeline:}
To address the necessity of answering macro-queries in the context of data exploration tasks; an investigation into the feasibility of adapting OpenAI's GPT models to generate charts from macro-queries was conducted through the development of an LLM pipeline.  While the primary focus pertains to macro-queries; the opportunity to address additional concerns was captured.  Specifically, due to the apparent lack of diversity in charts produced by other LLM systems \cite{tian2023chartgptleveragingllmsgenerate,dibia2023lida,10121440}; the decision to utilize Andrew Abela's Taxonomy\cite{abela} as guidance for selecting among an array of charts was made.  Moreover, Andrew Abela's Taxonomy\cite{abela} was chosen amongst others as it is widely cited\cite{10.1145/3543829.3544534}\cite{draux2020visualization}\cite{hellbruck2019analysis}\cite{granitto2022pictures}\cite{georgsson2018visualization}\cite{maguire2012taxonomy}\cite{munoz2017global}.

% evident

\begin{figure}[h]
    \centering
    \includegraphics[width=0.9\textwidth]{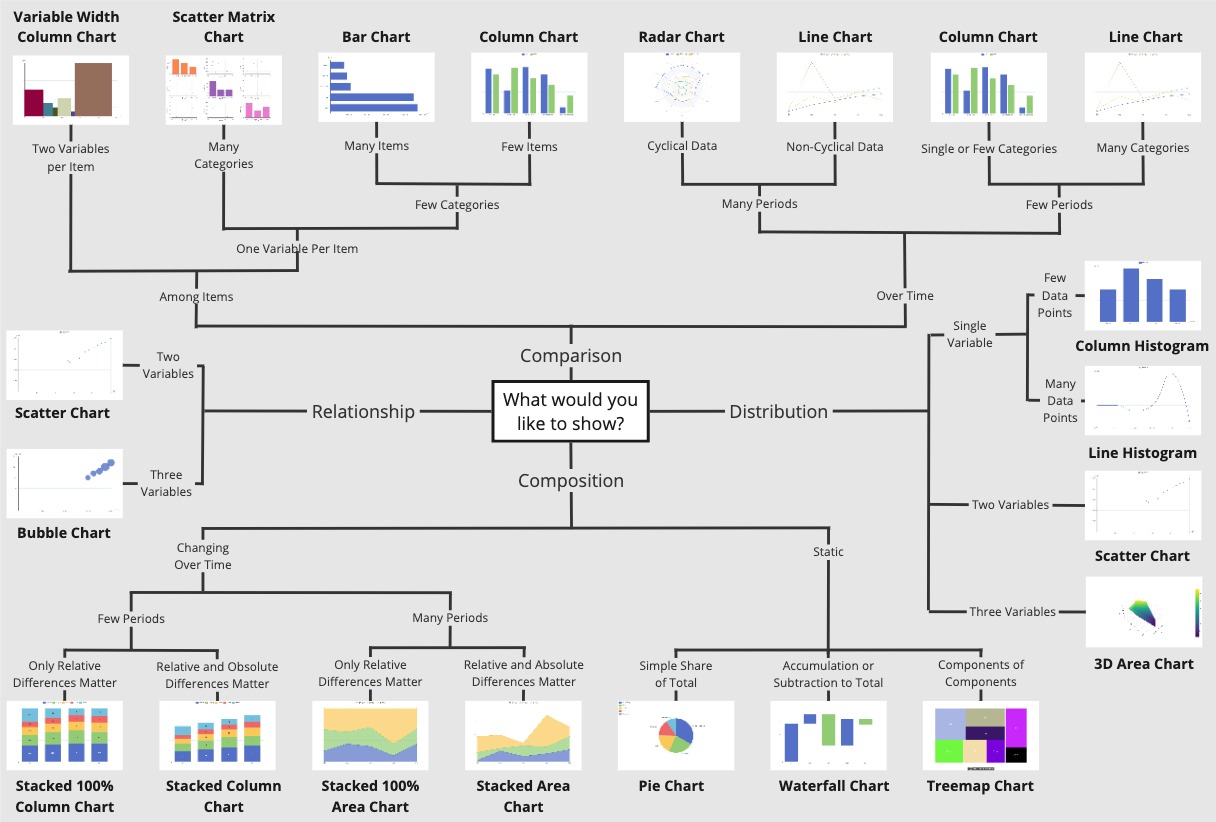}
    \caption{Andrew Abela's Chart Taxonomy\cite{abela} with chart templates superimposed and alteration from components of components to treemap chart since both charts embody equivalent principles.}
    \label{fig:tax}
\end{figure}

% all of the charts present in Andrew Abela's Chart Taxonomy was conducted.  Additionally, the (we also took the) opportunity because of the evident lack of diversity in default charts produced by LLM systems\cite{}.  We opted for a treemap opposed to components of components; as both charts embody equivalent principles.  Since there is an e.  We aim to develop a robust system that allows for explainable and accurate responses using a gamut of charts prompted by macro-queries.

%in an effort to utilized new developments in the academia and industry

%cannot be done intuitively or instinctively and requires  

% Specifically macro-queries entails a broad request to the LLM that does not involve directive utterances

% capable of ingesting large datasets,

% ... we want ...
% natural language high level queries
% transformations
% large datasets
% accuracy
% reasoning/ explain-ability
% no failure

\subsection{Contributions}

Our main contribution is the introduction of macro-queries, distinctly classified during the development of our prompt-to-visualization pipeline, which utilized Abela's Chart Taxonomy\cite{abela}.

 \begin{itemize}
  \item Macro-queries: A major aspect we hope to contribute is the introduction of the term macro-queries in context of data visualizations.  Moreover, we developed a pipeline to attempt to translate these macro-queries into actionable data exploration queries.
  \item Classification via LLMs with Abela's Chart Taxonomy\cite{abela} as reference: We demonstrate an attempt for some level of feasibility in leveraging LLM with Abela's Chart Taxonomy\cite{abela} to generate a diverse set of charts.
\end{itemize}

\section{Related Work}
\label{sec:others}
As research in LLM progresses rapidly, we observed similar work during our development phase. However, due to the fast pace of progress and the vast amount of information available, capturing the entire scope of comparable projects is unlikely~\cite{tian2023chartgptleveragingllmsgenerate,dibia2023lida,10121440}.

%There are some additional concerted works in the community to improve chart generation with artificial intelligence. \cite{}\cite{}

% Say here: there is a lot of prior work in translating natural language queries to visualizations - cite a synthesis paper (ask RJ which one). Then say, one of the earliest work among these is the foundational work we conducted in Articulate.
According to Shen \textit{et al.}\cite{Shen_2023}, there has been numerous prior work in natural language queries to visualizations and one of the earliest residing among them is Articulate\cite{5333099,sun2014articulate}.  As such our work builds upon our foundational concepts presented in Articulate\cite{5333099,sun2014articulate} involving the semi-automated generation of meaningful visualizations for non-experts using natural language,  which has since progressed into an always-listening natural language interface for creating data visualizations known as Articulate+\cite{10.1145/3543829.3544534,10.1145/3581641.3584079}. We envision that macro-queries will be another aspect which allow for vague and naturalistic interactions between humans and AIs.  Additionally, the opportunity for AI to spontaneously interject emerges when incorporating LLM due to its capability of interpreting indirect requests and macro-queries.

% Furthermore, observed discussions about visualizations in Artciulate+

% Additionally, utilizing LLMs allows for the interpretations of indirect requests and macro-queries to be processed and excecuted.

%With the anticipation of continued rapid developments surrounding LLM, specifically improved inference time, this research has the potential to be used as a substitution for specific elements of the NLP pipeline found within Articulate+.

%Ideally our proposed solution will be an evolution upon specific aspects of the NLP pipeline found in Articulate+\cite{}. 

% \subsection{Natural Language to Visualization}
Our work follows a similar guided approach to ChartGPT\cite{tian2023chartgptleveragingllmsgenerate}, however we aim to differentiate ourselves by incorporating a wider selection of charts and the usage of SQL queries to transform the data in accordance with  users' requests.

% \subsection{LLM with SQL}
There exists prior research into leveraging LLMs to translate user queries into SQL~\cite{10542806,10554719,10580129}.  However there is an exemplary lack of demonstrations associated among data visualization, with the recent exception of this Chat2Query\cite{10597681}.

% ChatGPT, chart consistency, using templates for consistency

\section{Architectural Approach}
\label{sec:headings}

\begin{figure}[h]
    \centering
    \includegraphics[width=0.9\textwidth]{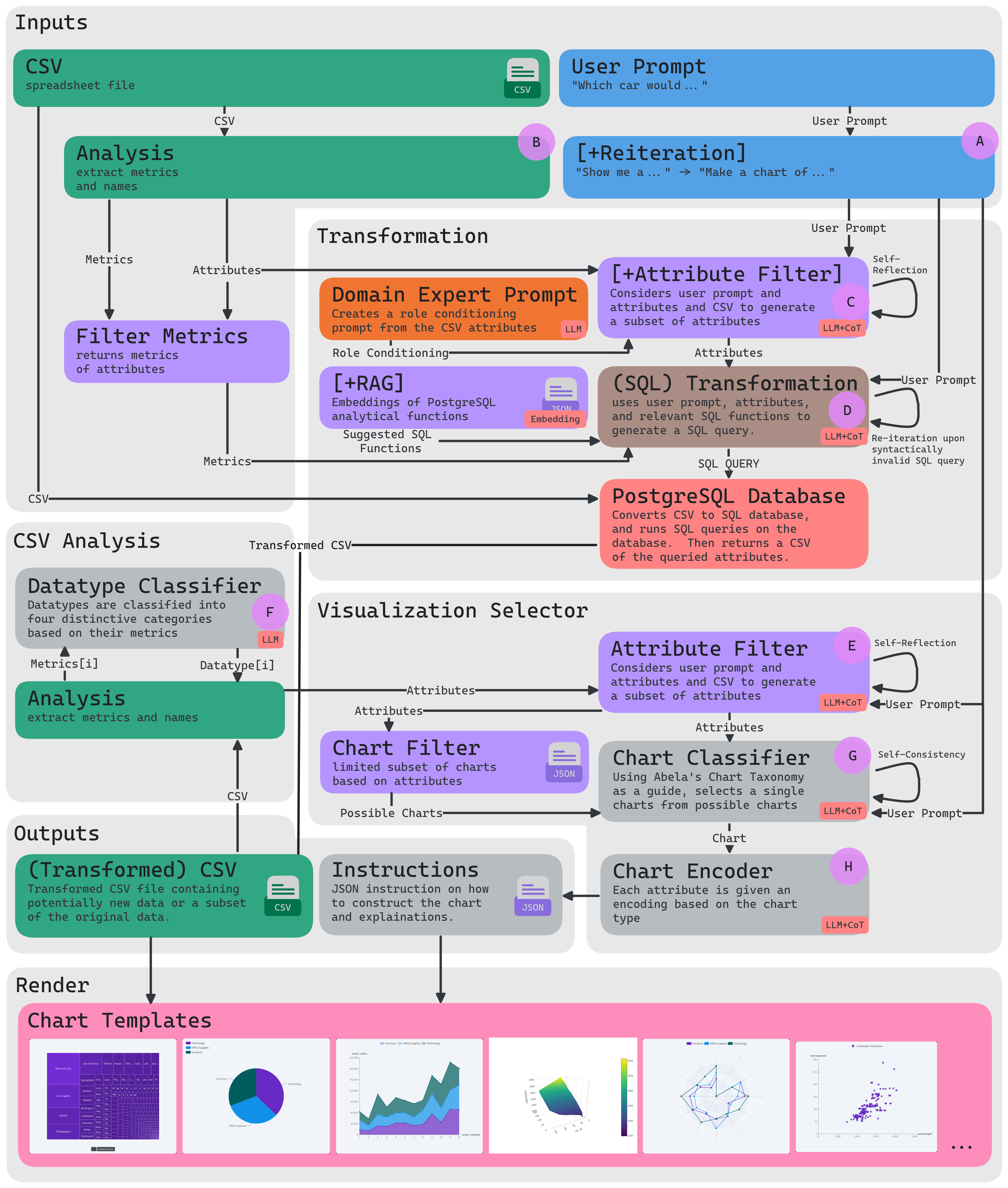}
    \caption{Model Architecture, where the inputs are a CSV and user prompt and the output is a JSON describing how to construct the visualization with the relevant CSV.}
    \label{fig:model}
\end{figure}

The architecture proposed incorporates prompt chaining for a multi-step guided approach with external tools to generate charts from a macro-query, and a CSV file containing relevant data,  relegating the LLM to a means of extracting and filtering critical information from a user's query and generating SQL queries.  

%This solution was built upon the extensive iteration of prior approaches conducted internally.

\subsection{Early Attempts}
The initial prototype utilized a decision tree-like approach (similar to that used in our prior work in Articulate and Articulate+\cite{5333099,sun2014articulate,10.1145/3543829.3544534,10.1145/3581641.3584079}), with traversal at each junction determined by the LLM.  Each split was hard-coded in accordance with Andrew Abela's Chart Taxonomy \cite{abela}.  Unfortunately this solution, when generating multiples of the same chart, produced inconsistent designs.  Furthermore, in the event that the user provided an ambiguous query, only one solution is presented.
% incongruent

In our second attempted prototype, we demonstrated the viability of fine-tuning and few shot prompting a LLM to favor a wider gamut of charts and partially align the model with Andrew Abela's Chart Taxonomy\cite{abela}.  The fine-tuned model is trained from data provided by a pilot survey conducted within our visualization lab.  The data, containing the desired result, and user prompt, was augmented using LLMs to include reasoning.  This fine-tuned system, given a user's prompt as input produced an array of possible charts due to imperfect information.  In our cursory testing, the resulting solution maintained a higher accuracy than a LLM following a decision tree-like approach.  An important insufficiency present in these iterations are the absence of transformation functionality for manipulation of data; inclusive of aggregations and filtering.

% REWRITEEEEEEEEEEEEEEEEEEEEEEEEEE
\subsection{Current Architecture}
% The final solution's feature set and characteristics were built upon the prior experiments.  

% Say explicitly: To address the limitations described in the previous section, Figure 2 shows our latest approach.

% This new architecture introduces a transformation module blah blah blah...

To address the aforementioned limitations,  Figure~\ref{fig:model} describes our latest approach.  This new architecture introduces a transformation module which performs text-to-SQL and utilizes chart templates to maintain consistent designs across multiple chart generation attempts.  Combining these separate modules in sequence involved prompt chaining which also provided more control over each individual step during the development phase.  Additionally, chain-of-thought\cite{wei2023chainofthoughtpromptingelicitsreasoning} was incorporated to improve performance and offer a degree of explainability.

% Subsequently, this has been resolved with the introduction of a transformation module performing text-to-SQL.  Furthermore, the final iteration offers a degree of explainability by explicating requesting chain-of-thought\cite{wei2023chainofthoughtpromptingelicitsreasoning} behaviour.  Adversely, as to not overload the LLM, the choice to develop an architecture with prompt chaining\cite{} provided more control over each individual step during the development phase.  

% Moreover, the LLM is essentially relegated to semantic filtering.

% Moreover, it proved beneficial in diminishing a majority of conversational exchanges arising from end user interventions or adjustments.  Thereby allow the user to effortlessly invoke a systematic chain of specially crafted prompts via one prompt.

% However, this choice may have been an artifact while working with less competent models such as Llama2 and its derived models\cite{}\cite{}\cite{}; therefore it may no longer be as necessary since migrating to GPT4 and GPT4o.
\subsection{Walk-through}
The input to the pipeline is a CSV file and a user prompt.  The user prompt should be relevant to the CSV.  Assuming a car based dataset, the user hypothetically could ask "What is the most affordable car?"  The prompt is passed to the optional reiteration step (\texttt{A} in Figure~\ref{fig:model}), which would rephrase the request.  Next, the CSV's headers are extracted and presented to the attribute filter (\texttt{C}) which will filter out a subset of attributes from entire set of attributes.  As an example, ["name", "price", "wheel base"] would become ["name", "price"].  This is passed to the SQL transformation (\texttt{D}), in which sorting by price would occur.  The results are then converted back into a CSV and attributes are filtered (\texttt{E}) once more, as it is possible that additional irrelevant attributes were created during the SQL transformation process.  Next, a list of all feasible charts are filtered after using chart filter which considers the datatype and attribute count.  Using a GPT4o-mini model fine-tuned on Abela's Taxonomy\cite{abela} (\texttt{G}), the system attempt to predict the ideal chart given the list of feasible charts.  Lastly, the encoder step  (\texttt{H}) ensures that the data types are in its proper encoding (e.g. x-axis, label, etc.).  This information and the transformed CSV from the transformation step  (\texttt{D}use the information to select the proper chart template and construct the visualization.

\subsection{Modules}
\paragraph{API and Webapp:}
For the prototype's implementation, the LLM was wrapped within an API where the inputs are the dataset, user prompt, and whether or not to return all feasible charts or AI recommended charts.  The returned values are the chart selections with attributes, modified dataset, and reasoning which are utilized by the fronted webapp(render).  This can be seen in Figures~\ref{fig:home}, ~\ref{fig:demosplit1} and \ref{fig:demosplit2}.

\paragraph{(Optional) Reiteration (\texttt{A} in Figure~\ref{fig:model}):} This optional step allows for ambiguous user requests, spoken utterances without explicit instructions, or other noisy data to be transformed into user command  A contrived example such as "I wonder what is the relationship of X and Y." should ideally be transformed into "Show me a visualization of the relationship of X and Y."  The implementation may vary on the use-cases and may not always be required.  Furthermore, this impromptu speech, without direct request for chart generation, aids in integration for an always listening system such as Articulate+\cite{10.1145/3543829.3544534}.

\paragraph{CSV Decomposition/ Analysis (\texttt{B} in Figure~\ref{fig:model}):}
This codified step involves a limited analysis on each attribute in a provided CSV file.  The following features are extracted per attribute: count, unique values, extremes, mean, standard deviation, variance, and top 5 values.  This limited feature set, a reduced subset of a more compressive list, was kept due to lack of evidence for significant quality variances in the LLM's responses.  This provided an excellent balance between inference times, and the ability to process larger datasets without hitting context lengths in future steps; effectively compressing the important features of the attributes.

\paragraph{(Optional) Attribute Filter (\texttt{C} in Figure~\ref{fig:model}):}
In this step, the main goal is to filter relevant attributes pertaining to the user's request from the entire list of attributes provided by the input CSV file.  Chain-of-thought\cite{wei2023chainofthoughtpromptingelicitsreasoning} is introduced in the pipeline to produce thoughtful responses with reasoning.  Furthermore, to ensure robust behaviour, the LLM's solution are verified by checking if attributes present in the response are a subset of all possible attributes found in the CSV.  In the event of failure, self-reflection of n times is used to encourage the LLM to resolve its error.  Lastly, role prompting, extracted via a LLM with the full set of attribute names as the input, is inserted into the context to steer the LLM to the relevant domain pertaining to the dataset.  We have observed benefits in regards to more applicable attribute selected and interpretation of vague and indirect user request when including role prompting.  These vague request are typically associated with macro-queries, that do not explicitly declare the desired attributes to analyze from the dataset.  This step is optional, as the SQL transformation step could fulfill a similar function.  However more testing is needed to verify the difference.

\paragraph{(SQL) Transformation (\texttt{D} in Figure~\ref{fig:model}):}
% The transformation prompt incorporates the careful arrangement, within the prompt, of: relevant attributes and the analysis performed filtered by the attribute limiter.

Utilizing a conversion from the CSV into a SQL database, we are able to provide transformations such as filtering, aggregation, sorting, and other prominent transformation capabilities provided by SQL via a crafted transformation prompt to the LLM.  This step provides the greatest impact in deciphering the user's macro-queries.  Specifically PostgreSQL was chosen due to its built-in analytical functions which allows for transformations such as linear regression or correlations.  By default, GPT4 fails to consistently utilize these addition transformations.  However, by restructuring functions descriptions found in the PostgreSQL documentation into JSON format and utilizing RAG (Retrieval-Augmented Generation)\cite{lewis2021retrievalaugmentedgenerationknowledgeintensivenlp}; the most relevant functions are extracted based on the user's prompt and appended to the transformation prompt's context.  As a fail safe, we verify that the SQL query generated by the LLM is valid and reiterate up to a user-defined n amount of retries.  The choice for a re-attempt (running the prompt again) instead of self-reflection was done after noticing that, in our limited testing, self-reflection, typically triggered on more challenging requests, continuously failed due to small irrelevant modifications of its initial SQL query rather than a redo.  This can be associated with the degeneration-of-thought problem\cite{liang2024encouragingdivergentthinkinglarge}.  In the rare occurrence that more than n equals four retries fails, a fallback solution is provided to bypass the transformation step. Thereafter, SQL response is converted into a CSV format to be analyzed further down the pipeline.  We decided to explore SQL for transformations due to favoring a more restrictive scope against programming languages such as python and for potential future implementation with big data.  Lastly, if the resulting SQL response contains one row, excluding headers, the system shall return a table.  This is due to the fact that charts containing one data point are not helpful.
%  and have methods to notify the user in the final result.

\paragraph{(Charting) Attribute Filter (\texttt{E} in Figure~\ref{fig:model}):}
Similar to the prior attribute limiter, another attribute limiting step is introduced with explicit directives to favor two to three important attributes while maintaining attributes critical to human interpretation (e.g. unique identifiers, names, etc.).  This additional step is crucial in counteracting irrelevant, additionally generated, attributes from the transformation phase.  Additional attributes may arise when prompting the LLM with "best bang for buck", "averages", etc.

\paragraph{Datatype Classifier (\texttt{F} in Figure~\ref{fig:model}):}
This step tasks the LLM to infer one of the four datatype (nominal, ordinal, discrete, continuous) based on analysis performed on the transformed CSV.  With our limited testing, omission of chain-of-thought\cite{wei2023chainofthoughtpromptingelicitsreasoning} had minimal bearing on the accuracy of the final result.

\paragraph{Chart Classifier (\texttt{G} in Figure~\ref{fig:model}):}
Chart selection is accomplished by searching the space of all possible predefined chart templates available in the system and selecting all syntactically valid charts based on attribute counts and attribute types.  Effectively presenting a constraint satisfaction solution.  For this step, either one of the two solutions can be provided via a conditional statement.  The first returning the entire subset of feasible charts.  The second incorporates a fine-tuned LLM, trained from a vetted LLM generated datasets with influence from Abela's Chart Taxonomy\cite{abela}, and chain-of-thought prompting to determine the most appropriate chart type from the given set of all possible chart types.  
% The commented out is no longer true
% The absence of the user prompt produced better results as it removed confusion generated by initially declared user attributes and the post transformation attributes.

\paragraph{Chart Encoder (\texttt{H} in Figure~\ref{fig:model}):}
In the event that the LLM determines the chart type, it will also provided the encoding associated.  This is done using instructions pertaining to chart type with associated encodings and JSON template in which the LLM assigns the optimal attribute to the encoding such as axis, which attribute should be binned, attribute for frequency, etc.

\paragraph{(Render) Chart Templates:}
The rendering of the charts is done via a webapp on the front end.  The inputs are the transformed csv and the chart selection with encoding.  This information is fed into a chart tempting system and produces the chart.
% We use chain-of-thought\cite{wei2023chainofthoughtpromptingelicitsreasoning} and re-iteration\cite{} with table as a fallback.  However, it may prove invaluable to incorporate fine-tuning as this task may require deeper domain knowledge.  There is no strict adherence to the Abela's Taxonomy, however each chart's descriptor has been properly inspired by Abela's Chart Taxonomy\cite{}.

% \paragraph{Chart Generation:}
% The rendering of the charts is done on the front end

%The chart generation module takes the output from the prior pipeline to generate charts and provide users with an explanation at each stage. The output consists of: explanations at each stage, a new CSV file generated through the SQL transformations, and the chart type. The chart generation module primarily utilizes the generated CSV file and chart type to create a chart. Chart attributes are selected at random, however, in the future, there can be guided constraints on where attributes should be. The model’s output is designed to be charting library agnostic, to ensure maximum flexibility. Connecting this prompt-chained LLM to a chart generating agent could be useful in the future.

% (maybe could explain why ChatGPT can’t do it all, worth talking about needle in the haystack tests and so forth?)

% We define macroqueries as high-level queries where the LLM (Large Language Model) uses its training data to create relationships with the new data. An example of this is, <insert example here, top 10 cars in the US, most bang for the buck, etc.>.

% Make mention of template system
% List of charts we can generateted

\begin{figure}[!tbp]
  \centering
  \begin{minipage}[b]{1\textwidth}
    \includegraphics[width=\textwidth]{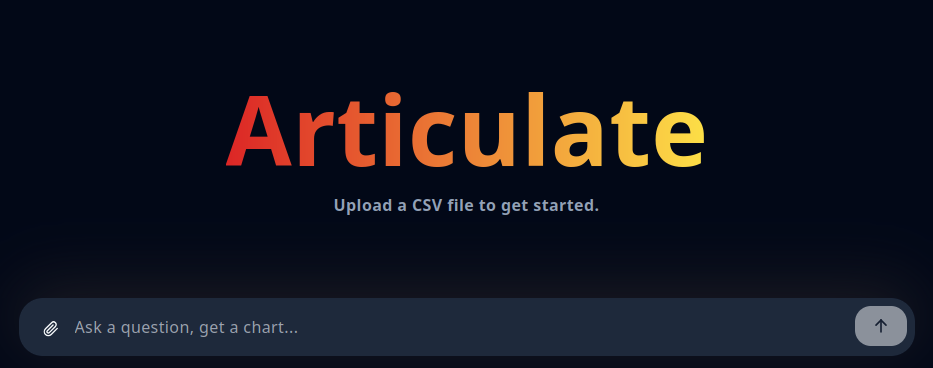}
    \caption{Home page of the web app utilizing the API}
    \label{fig:home}
  \end{minipage}
\end{figure}

\begin{figure}[!tbp]
  \centering
  \begin{minipage}[b]{0.48\textwidth}
    \includegraphics[width=\textwidth]{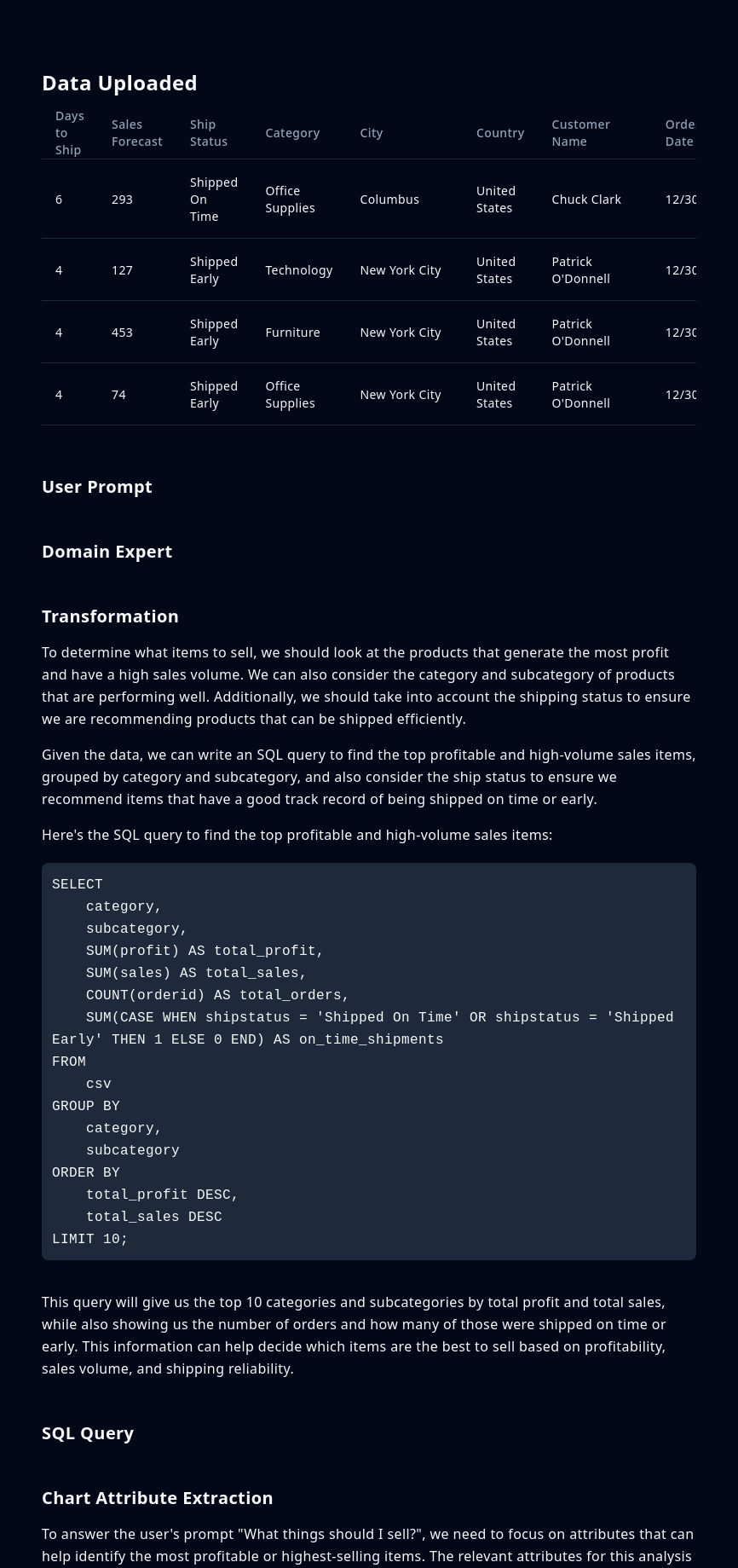}
    \caption{Response with reasoning at each step if applicable invoked by the macro-query: "What things should I sell?"}
    \label{fig:demosplit1}
  \end{minipage}
  \hfill
  \begin{minipage}[b]{0.48\textwidth}
    \includegraphics[width=\textwidth]{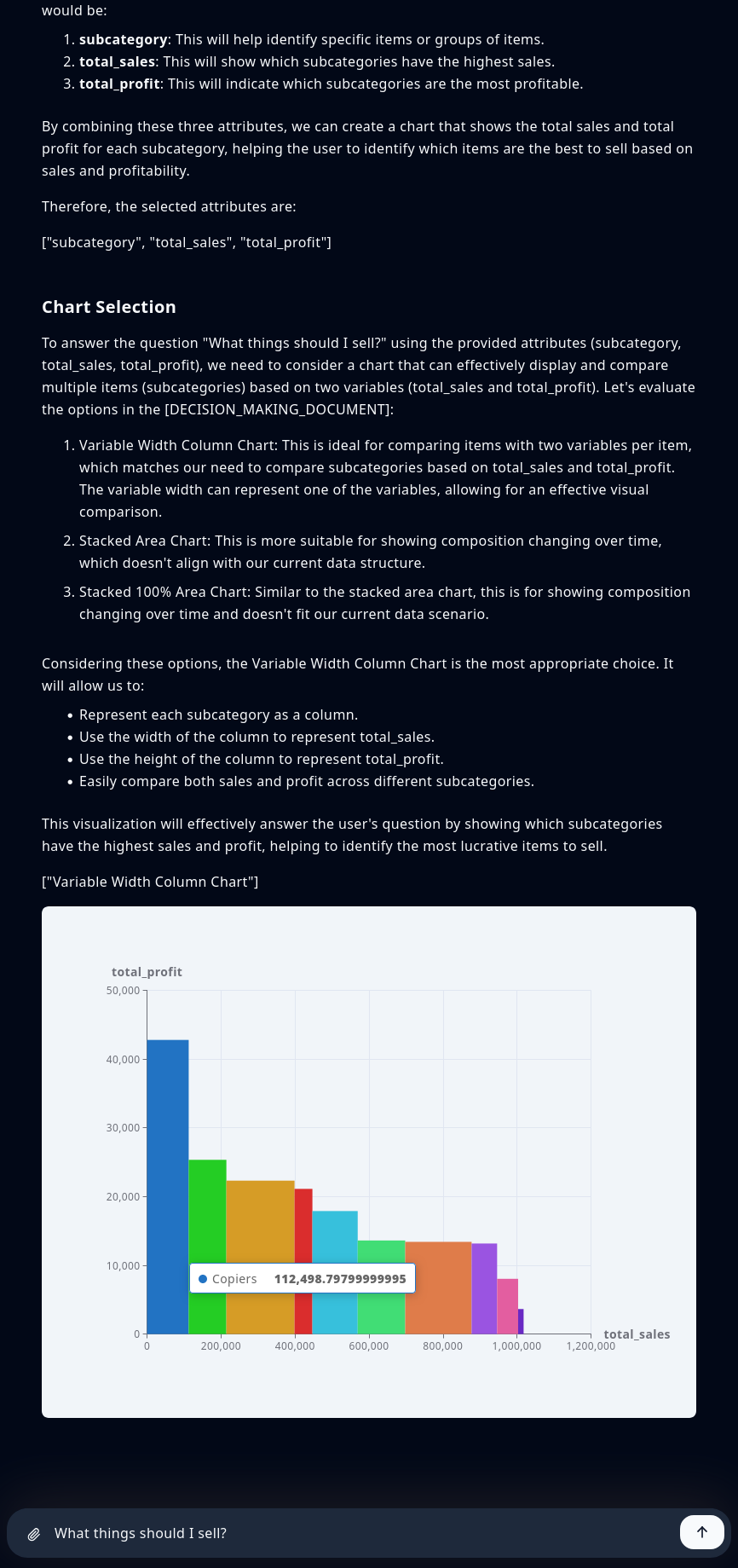}
    \caption{Continuation of Figure~\ref{fig:demosplit1}'s reasoning with a interactive variable width column chart generated based on Abela's Chart Taxonomy}
    \label{fig:demosplit2}
  \end{minipage}
\end{figure}

% \begin{figure}[!tbp]
%   \centering
%   \begin{minipage}[b]{0.32\textwidth}
%     \includegraphics[width=\textwidth]{articulatehomepage.png}
%     \caption{Home page of the web app utilizing the API}
%     \label{fig:home}
%   \end{minipage}
%   \hfill
%   \begin{minipage}[b]{0.32\textwidth}
%     \includegraphics[width=\textwidth]{Capture-2024-07-15-091410.png}
%     \caption{SQL query displayed amongst a sea of reasoning}
%     \label{fig:sql}
%   \end{minipage}
%   \hfill
%   \begin{minipage}[b]{0.32\textwidth}
%     \includegraphics[width=\textwidth]{Capture-2024-07-15-091322.png}
%     \caption{Generated chart from a user's prompt}
%     \label{fig:chart}
%   \end{minipage}
% \end{figure}

%In the event that the API returns

\subsection{Design Rationale}

\paragraph{LLM Choice:} While the field of artificial intelligence rapidly progresses, this modular approach promotes interchangeability of LLMs at each stage.  Thereby allowing an improvement in specific regions such as attribute selection or SQL generation given that a superior model exists.  As for this research, the utilization of LLMs from OpenAI was due to their state of the art nature.

\paragraph{GPT4 vs GPT4o in Transformations:}While developing this software; we noticed that the transformation step performed better using GPT4 instead of GPT4o.  The responses from GPT4 for SQL typically included a more thoughtful response when both prompted using chain-of-thought\cite{wei2023chainofthoughtpromptingelicitsreasoning}.  As an example, while asking "Which car would batman drive?", 

GPT4o produces the following SQL:

\texttt{
% \begin{lstlisting}[language=SQL]
    SELECT carname, horsepower, enginesize, price 
    FROM csv ORDER BY horsepower DESC, enginesize DESC, price DESC 
    LIMIT 1;
% \end{lstlisting}
}

However GPT4 responds (with comments omitted):

\texttt{
% \begin{lstlisting}[language=SQL]
    SELECT carname, horsepower, carbody, price FROM csv 
    WHERE (horsepower >= (SELECT MAX(horsepower) * 0.75 FROM csv))
    AND (carbody IN ('convertible', 'hardtop'))
    AND (price >= (SELECT MAX(price) * 0.75 FROM csv))
    ORDER BY horsepower DESC, price DESC;
% \end{lstlisting}
}

We also noticed that the transformation step will also filter out some unwanted variables, there may be potential to improve performance by focusing on fine tuning with SQL queries and removing redundant attribute limiter steps. 

\paragraph{Retrieval-Augmented Generation (RAG):}Regarding the utilization of RAG\cite{lewis2021retrievalaugmentedgenerationknowledgeintensivenlp} in the transformation step, since it is always selecting the top fifteen closest matches, there maybe notions of adverse effects.  However, during our testing, we did not notice any abnormalities that suspects the additional information provided by the RAG\cite{lewis2021retrievalaugmentedgenerationknowledgeintensivenlp} negatively impacts the desired outcome.  With the exception that some prompts trigger these functions and produces a single rowed spread sheet response which is unfit for graphing.  This is the case in the results Table~\ref{table:results3} where the temporary prompt injection solution to disable the SQL suggestions were needed.

\paragraph{Fallback:}Upon the LLM's failure to produce the desired response, the default choice is to skip the step instead of notifying the user of the failure.  However this rare occurrence could potentially lead to user confusion.  Alternatively, it may be ideal to halt the system instead of bypassing the step as this provides a harsher indication of an error and benefits accuracy over reliably.

Considering the chart selection module, the necessity of LLMs to invoke a single desired chart may be debatable.  As such it is not unreasonable to bypass this stage and allow the user to select from a range of possible charts with the LLM offering a suggested optimal.

\paragraph{Closure:}We did not explore the full gamut of prompting techniques and supplied the model with what we felt was necessary to achieve a functionally coherent solution.  Therefore, this may not be the most optimal solution available.
% When solely relying prompting techniques we noticed the the LLM typically favored a limited selection of common charts, even when presented with additional options.

\label{sec:headings}
\section{Preliminary Evaluation Methodology}
The following presents two preliminary tests to demonstrate the results of our system.  Theses evaluations are cursory and not conclusive of performance of all various types of datasets.

\subsection{(Future) Macro-Query Evaluation}
To verify the preliminary macro-query performance of our pipeline, four reviewers, collectively, evaluated responses provided by the LLM.  For each prompt, the model was using Car Price dataset (dataset representing a limited variety of cars and their associated featured such as mpg, brands, fuel type, door count, etc.)\cite{carprice} and a superstore dataset (days to ship, sales, profits, city, country, category, etc.)\cite{superstore}.  The only modification to the superstore dataset involved the removal the product name column that caused errors in the CSV ingestion step.  Our evaluation is subject to human approval; For as long as the LLM produces a chart that aided in answering the user's prompt, it is considered satisfactory.  Additionally, for the preliminary testing the chart encoder step was disabled due to implementation issues.

% To verify the preliminary performance of our pipeline, four reviewers, collectively, evaluated responses provided by the LLM.  For each prompt, the model was queried three times per prompts using Car Price dataset (dataset representing a limited variety of cars and their associated featured such as mpg, brands, fuel type, door count, etc.)\cite{carprice}.  Our rudimentary evaluation mainly involves macro-queries and is subject to human approval; For as long as the LLM produces a chart that aided in answering the user's prompt, it is considered satisfactory.
% for as long as the LLM does not produce any logical fallacies between the SQL and reasoning associated, it is considered satisfactory.  Additionally, the chart provided must aid in answering the user's prompt.  Therefore, this evaluation does not place any value on the most optimal response.

\subsection{Chart Diversity Evaluation}
Secondly, to verify that nearly all of Abela's charts\cite{abela} could be generated, we provided our system with a handcrafted golden set of prompts that followed Abela's logic and verified that the correct charts were produced.  To aid with testing, we opted not to utilize macro-queries in the prompts as their innate characteristics make them nondeterministic due to the wide spread of possible interpretations from the LLM.  Keywords from Abela's taxonomy was use to clarify intent due to plausible subjective interpretations of distribution, comparison, composition, and relationship from prompts if left unspecified.  In regards to fine-tuning, data typically included mention of car based dataset, testing was done using a modified superstore dataset\cite{superstore}.  The only modification being the removal the product name column that caused errors in the CSV ingestion step.  Additionally, for the preliminary testing the chart encoder step was disabled due to implementation issues.

% To aid in testing, specific Abela's language was use to clarify intent due to the distribution, comparison, composition, and relationship being subjective in nature.

% Which could be inclusive of a majority attributes, and multiple derived calculations.
% Therefore, each module's reasoning is subject to the evaluator's definition of coherency.  
%  in accordance to human approval.
% Completeness Evaluation: possbilly generated charts

\label{sec:headings}
\section{Discussion}

% During our tests with the model, we observed that it can establish connections between macro-queries and the provided data. 

% No Longer Valid Observation vvvvv
% During the evaluation phase; we discovered that while the transformation step exhibited SQL queries that addresses the user's request, the chart selection step showed a bias was shown towards more commonplace charts. Based on our prior efforts in fine-tuning LLM's to favor a wider distribution of charts, the same technique may prove invaluable in aligning the LLM to reduce its bias towards common charts such as scatter plots or bar charts.  Typically if the results are not as anticipated, the LLM's given reasoning may convince the user of alternative acceptable solutions.  This is commonly the case with opinionated macro-queries such as requesting optimal vehicles for mail delivery or camping.  Notably, the transformation stage encourages some form of formulated logic from.
%  movies to watch while sad
% ^^^^^^^^^^^^^^^

\label{sec:headings}
\subsection{Results}
The results indicate that additional modifications to the system are needed to improve the system for other various datasets.  More fine tuning and architecture work is needed to fully support and select all of Abela's Charts\cite{abela}.  However, we believe that the current results indicate potential and promise.
% Overall, we believe that we are on the right track.
\subsubsection{Macro-query Results}
% Our preliminary review has seen that a majority of the macro-query prompts are satisfactory.  However it shall be noted that formal review and results will eventually be conducted at a later time.
The following are a few examples of the results produced in our macro-query preliminary testing.  Table~\ref{table:macroresults} demonstrates the variability in interpretation given the same macro-query and Table~\ref{table:macroresultscar} demonstrates the LLM's various capabilities for inferring data and deriving metrics.

Regarding Table~\ref{table:macroresults}, The associated visualization answers the question, but typically with an oddity associated (same labels, different numbers).  This is because the transformation step took into consider the subcategories, categories, and segment during its calculations, however, some attributes are disregarded post transformation.  This can result in the chart generated to seemingly contain duplicate entries.  Which may be confusing for the end user.

In Table~\ref{table:macroresultscar}, regarding "what are the most bang for the buck european cars?" notice how the data augmentation occurs in addition to a LLM derived value score.  Resulting in an acceptable chart visualizing the spectrum of the highest scored vehicle.  This produced chart contains a cursor hover-over interaction denoting the vehicle name.  In the case for Batman; the LLM transcribes features associated with Batman to the relevant SQL query and generates a suitable variable width column chart featuring all potential car candidates.  It should be also noted that the pipeline is not perfect and can result in incomplete answers, such is the case for "What is the best car for camping?"  The results show a chart whose answer is not particularly useful and does not pertain directly to the question.  Lastly, regarding "fastest 0 to 60", the LLM derived a substitute metric due lack of information.  This produced a decent chart, however due to the extensive list of results, a user may only desire to view the top ten results when it comes to categorical attributes.  

In summary, the macro-query responses are not perfect, but are mostly acceptable.  The charts produced still require a minor adjustments to improve legibility.

% In table\ref{table:macroresultscar}, regarding "what are the most bang for the buck european cars?" notice how the data augmentation occurs in addition to a LLM derived value score.  This shows the potential to perform queries slightly outside the scope of the dataset.  Moreover, in the case for Batman; the LLM transcribes features associated with Batman to the relevant SQL query.  It should be also noted that the pipeline is not perfect and can result in incomplete answers, such is the case for "What is the best car for camping?"  The results show a chart whose answer is not particularly useful and does not pertain directly to the question.

\begin{table}
\caption{Generated Charts from Macro-queries.  Due to exorbitant word counts, only the transformation justification is shown.  Using the superstore dataset\cite{superstore} (with names column removed).}
\label{table:macroresults}
\begin{tabularx}{1\textwidth} {|p{0.15\textwidth}|p{0.80\textwidth}|}
 \hline
 \begin{center} \bf Prompt \end{center} & \begin{center} \bf Transformation Step Response and Visualization \end{center}\\
 \hline
what will make me the most money?  & \includegraphics[width=0.5\linewidth]{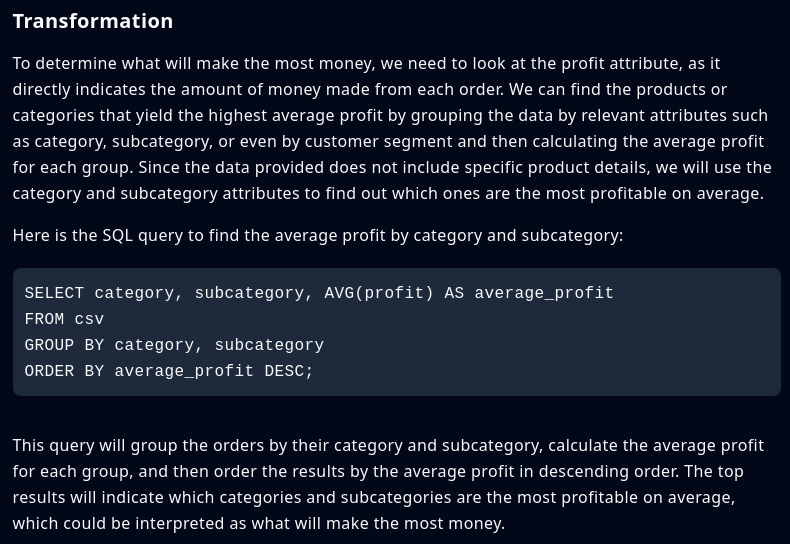}  \includegraphics[width=0.5\linewidth]{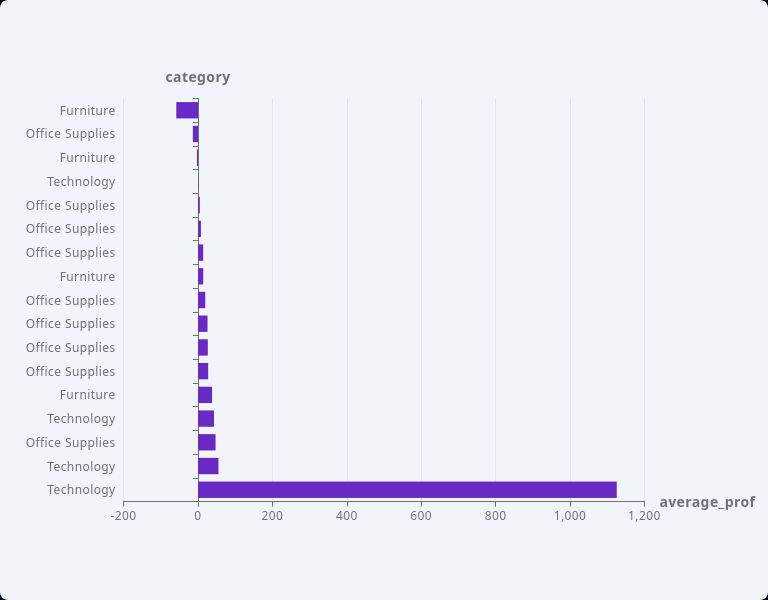}\\

\hline
what will make me the most money?  & \includegraphics[width=0.5\linewidth]{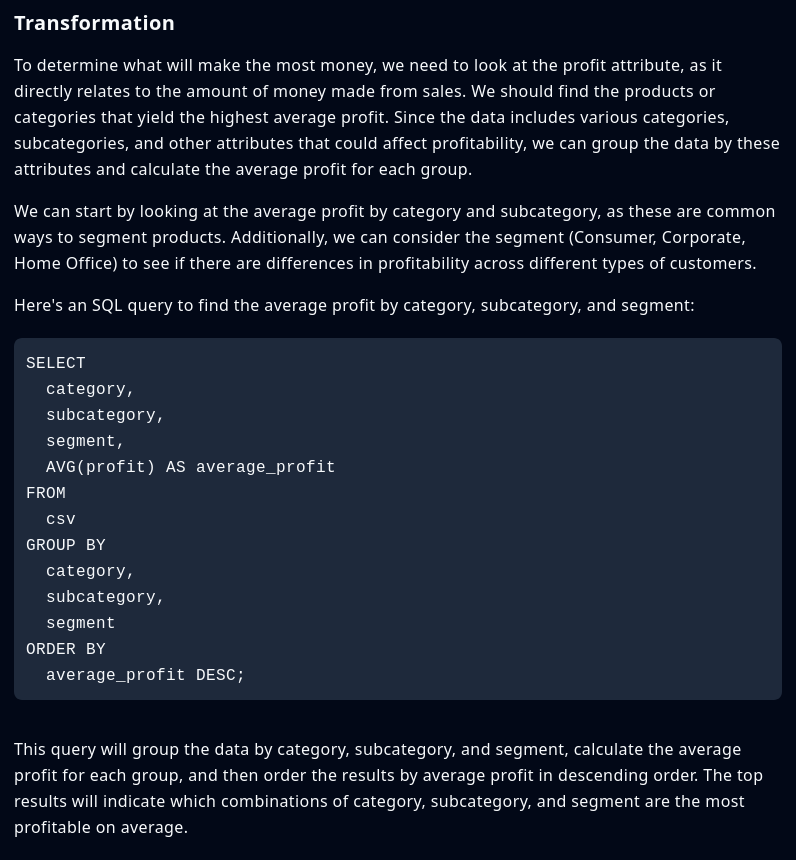}  \includegraphics[width=0.5\linewidth]{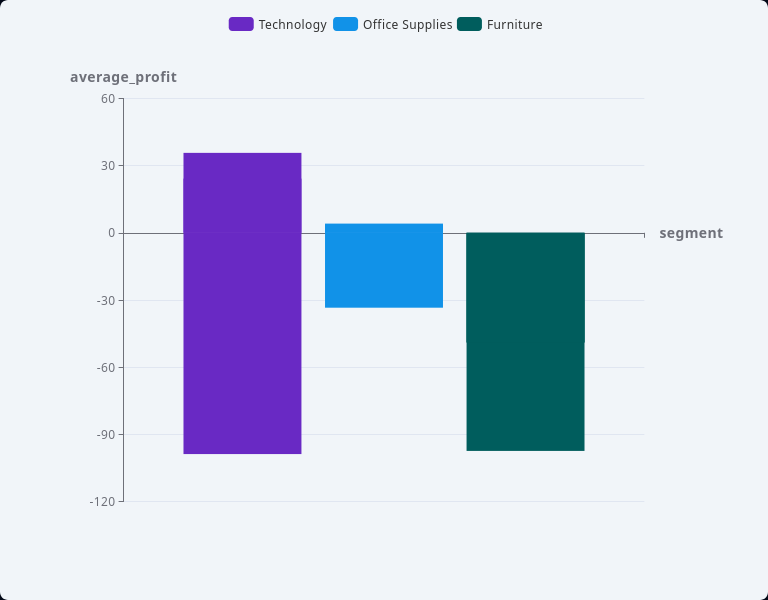}\\

\hline
what will make me the most money?  & \includegraphics[width=0.5\linewidth]{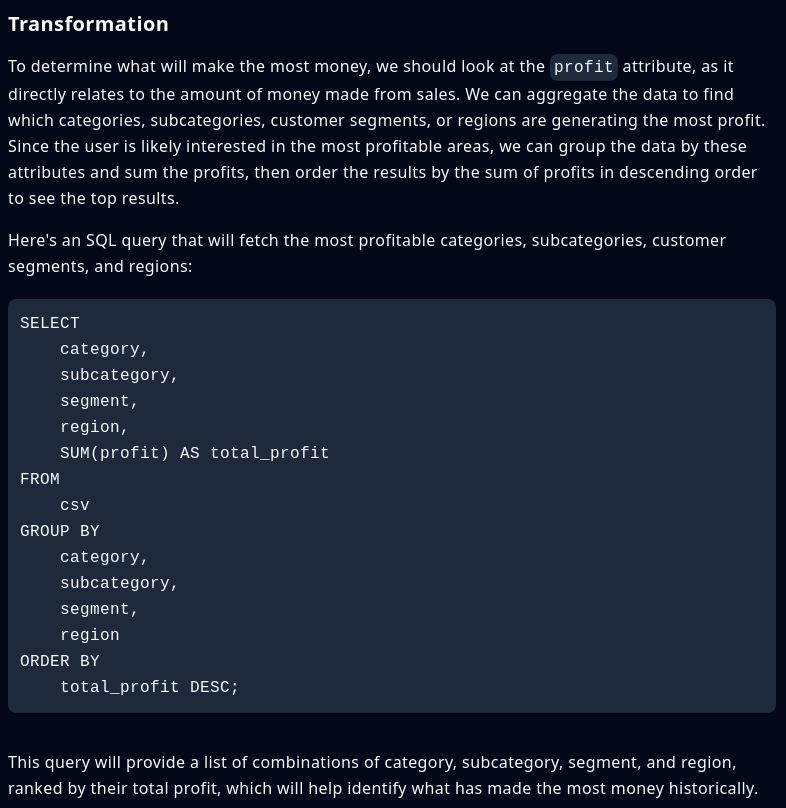} \includegraphics[width=0.5\linewidth]{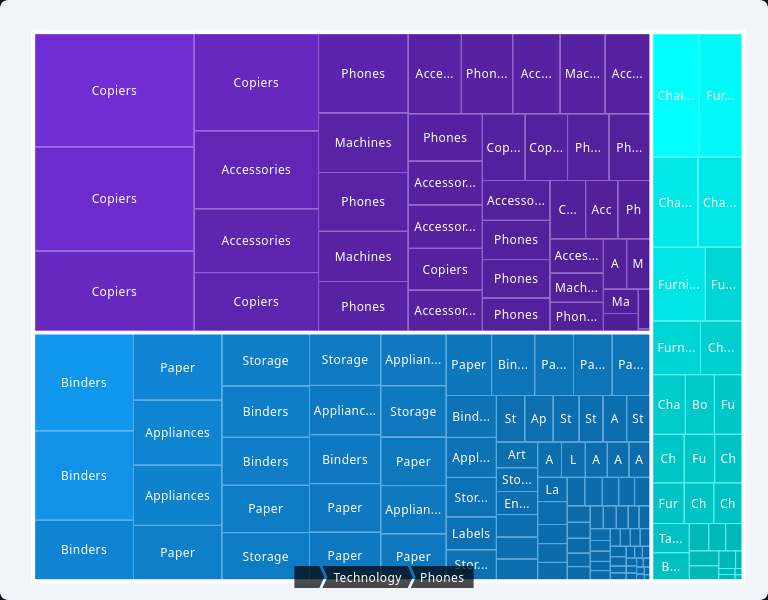} \\

% \hline
% what will make me the most money?  & \includegraphics[width=1\linewidth]{macro-results/Screenshot 2024-08-08 at 16-01-21 Create Next App.png} \hline \includegraphics[width=1\linewidth]{macro-results/Screenshot 2024-08-08 at 16-01-21 Create Next App.png}\\
\hline
\end{tabularx}
\end{table}

\begin{table}
\caption{Generated Charts from Macro-queries.  Due to exorbitant word counts, only the transformation justification is shown.  Using the Car Price dataset\cite{carprice}.}
\label{table:macroresultscar}
\begin{tabularx}{1\textwidth} {|p{0.15\textwidth}|p{0.80\textwidth}|}
 \hline
 \begin{center} \bf Prompt \end{center} & \begin{center} \bf Transformation Step Response and Visualization \end{center}\\
 \hline
what are the most bang for the buck european cars?  & \includegraphics[width=0.5\linewidth]{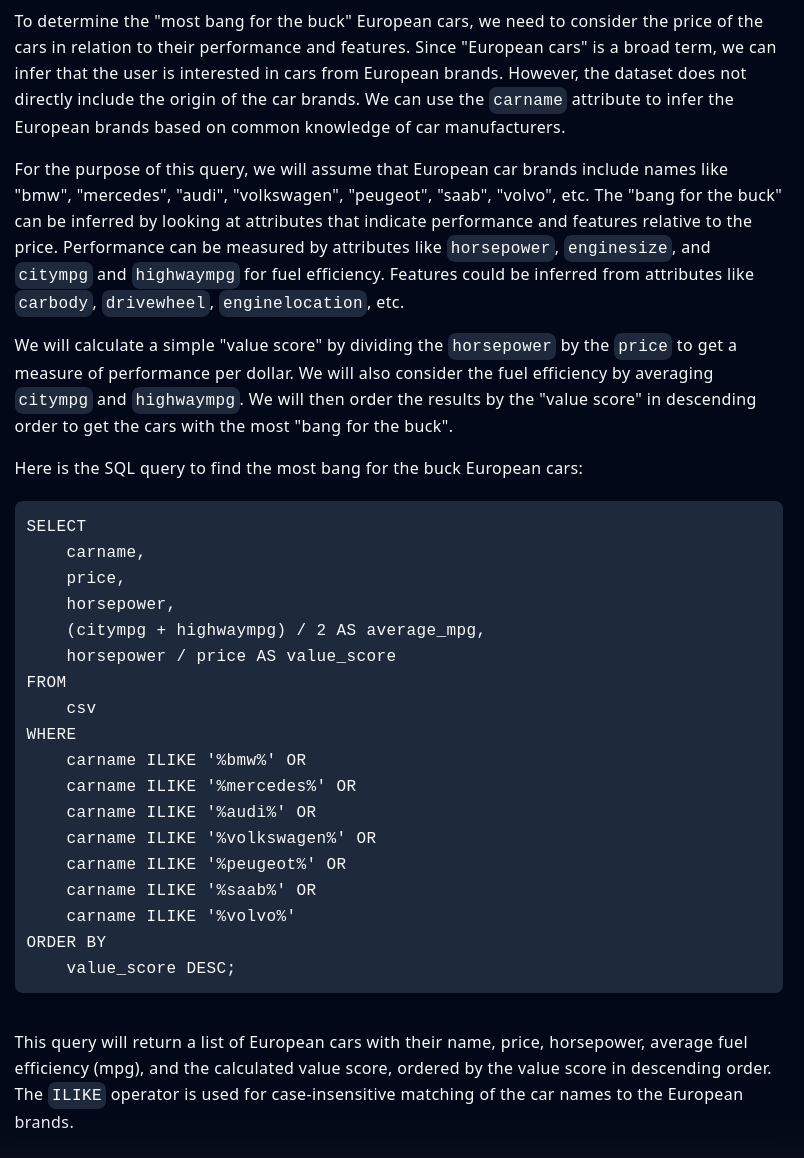} \includegraphics[width=0.5\linewidth]{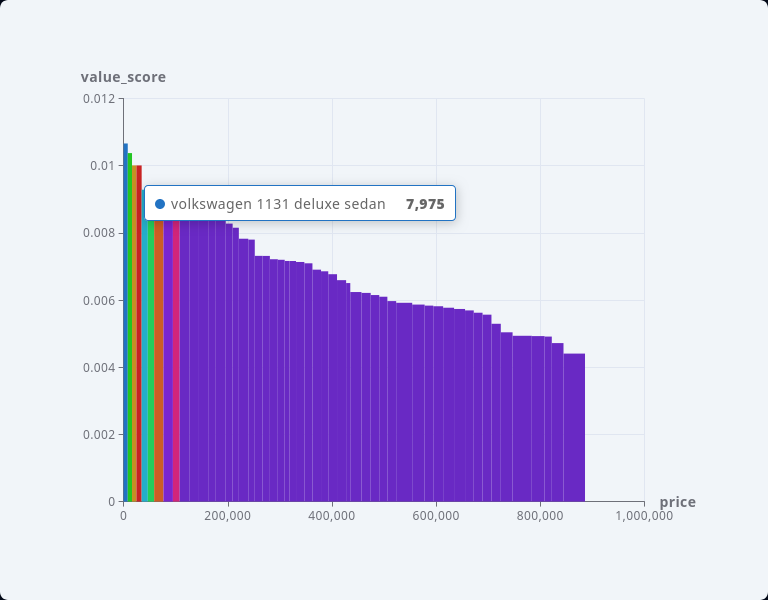} \\

 \hline
which car would batman drive?  & \includegraphics[width=0.5\linewidth]{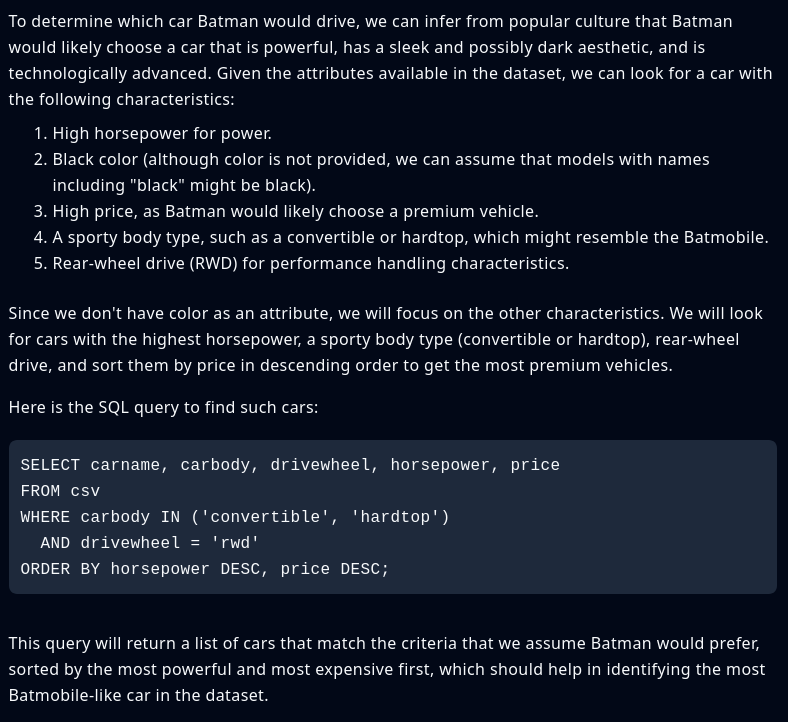} \includegraphics[width=0.5\linewidth]{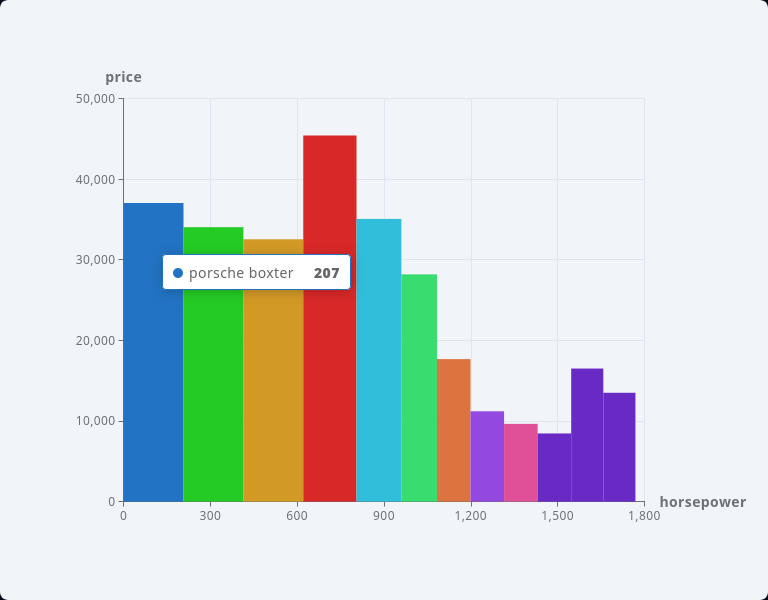} \\

 \hline
\end{tabularx}
\end{table}

\begin{table}
\caption{Generated Charts from Macro-queries.  Due to exurbanite word counts, only the transformation justification is shown.  Using the Car Price dataset\cite{carprice}.}
\label{table:macroresultscar2}
\begin{tabularx}{1\textwidth} {|p{0.15\textwidth}|p{0.80\textwidth}|}

 \hline
What is the best car for camping?  & \includegraphics[width=0.5\linewidth]{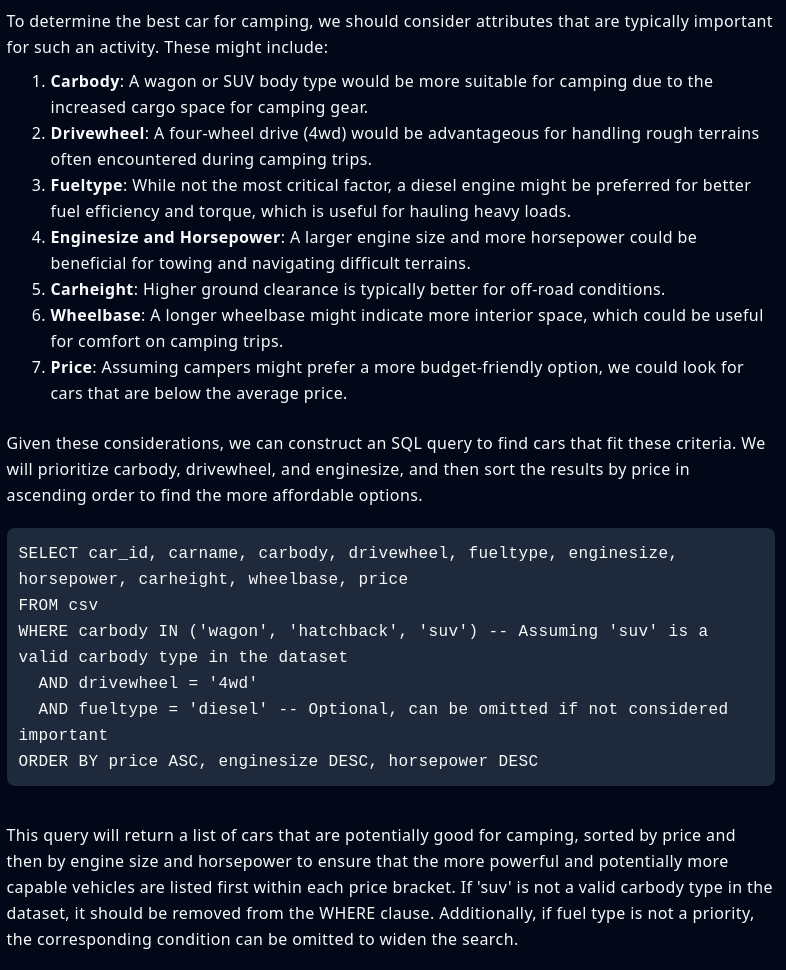} \includegraphics[width=0.5\linewidth]{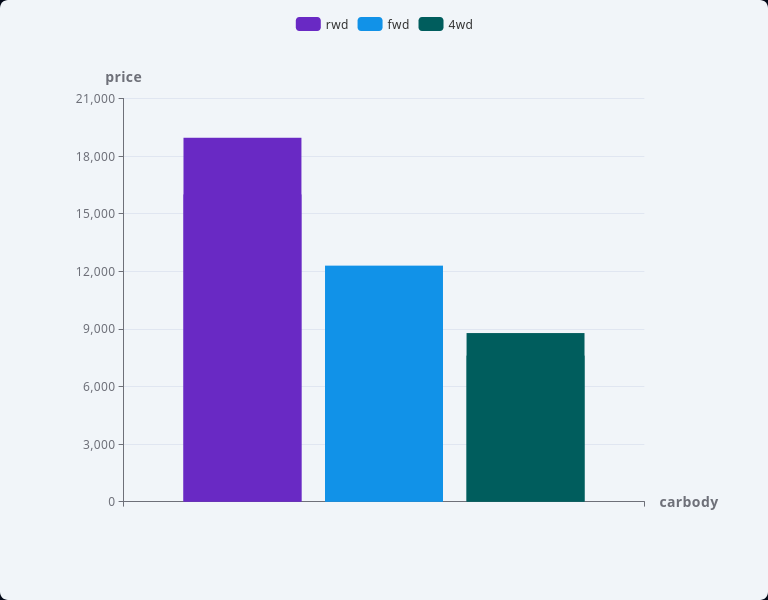} \\

 \hline
Which car has the fastest 0 to 60?  & \includegraphics[width=0.5\linewidth]{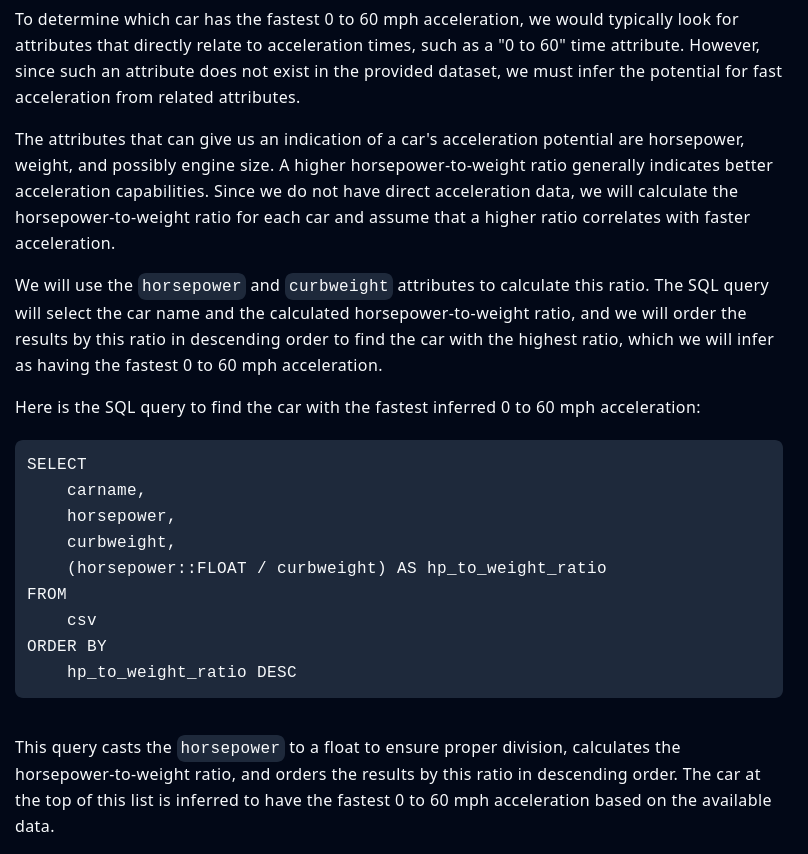} \includegraphics[width=0.5\linewidth]{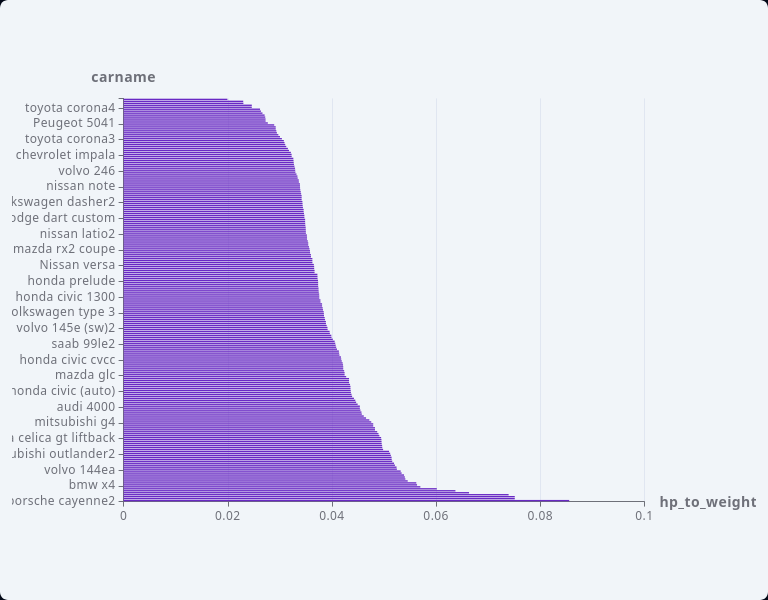} \\

\hline
\end{tabularx}
\end{table}

\subsubsection{Chart Diversity Results}
For the majority of cases in Tables~\ref{table:results1}, \ref{table:results2}, \ref{table:results3}, \ref{table:results4} and \ref{table:results5}; entering the prompt results in the visualization generated without fault.  However, we occasionally noticed that certain prompts had to be re-attempted to generate the correct chart.  

Regarding problems during inference, there were three distinct instances where the specific chart type must be stated to generate the chart.  Moreover, prompt injection in Table~\ref{table:results3} was required to generate the desired chart.  Lastly, We were unable to generate a scatter matrix chart due to the limitations of our approach; specifically the scatter matrix chart takes in a dynamic amount of attribute types and our system only works with a fixed attribute count.  This indicates that more work should to be done in this area.  Particularly, it may prove more beneficial, for accuracy, to manually create the training dataset rather than relying on LLM generation.

\begin{table}
\caption{Generated charts using the LLM pipeline}
\label{table:results1}
\begin{tabularx}{1\textwidth} { 
  | >{\raggedright\arraybackslash}X 
  | >{\raggedright\arraybackslash}X 
  | >{\raggedright\arraybackslash}X | }
 \hline
\begin{center} \bf Prompt \end{center} & \begin{center} \bf Chart Name \end{center} & \begin{center} \bf Visualization \end{center} \\
\hline
Show me a comparison of categories' quantity and sales forecast & Variable Width Column & \includegraphics[width=0.31\textwidth]{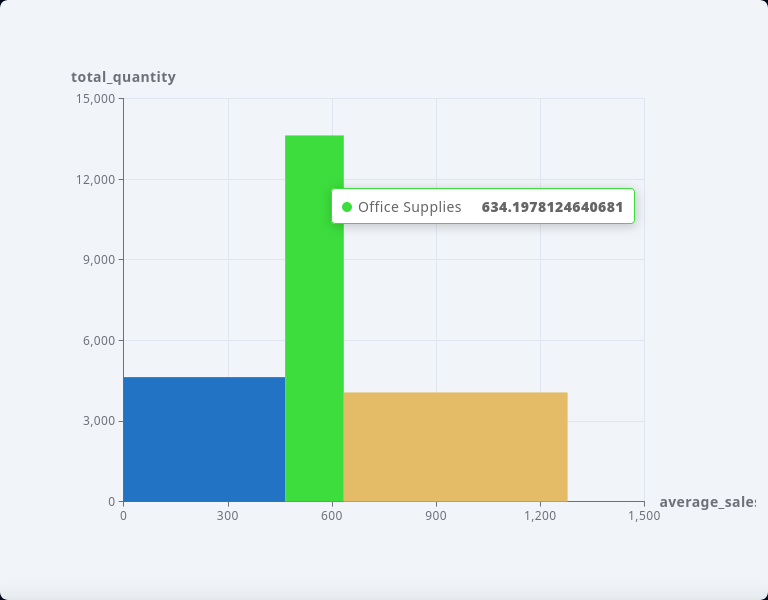}\\
\hline
(Skipped, unable to generate) & Scatter Matrix Chart & N/A\\
\hline
Show me a comparison of the count of cities per category & Bar Chart & \includegraphics[width=0.31\textwidth]{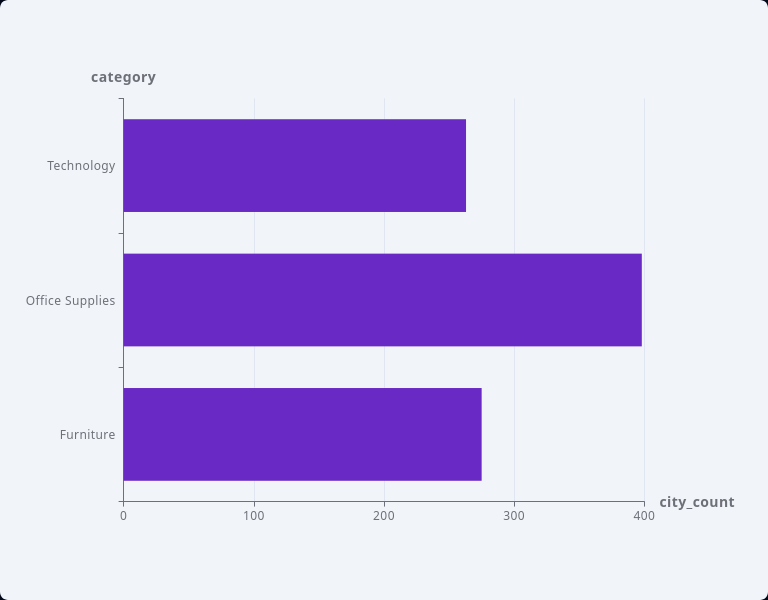}\\
 \hline
Show me a comparison of segment per category & Column Chart & \includegraphics[width=0.31\textwidth]{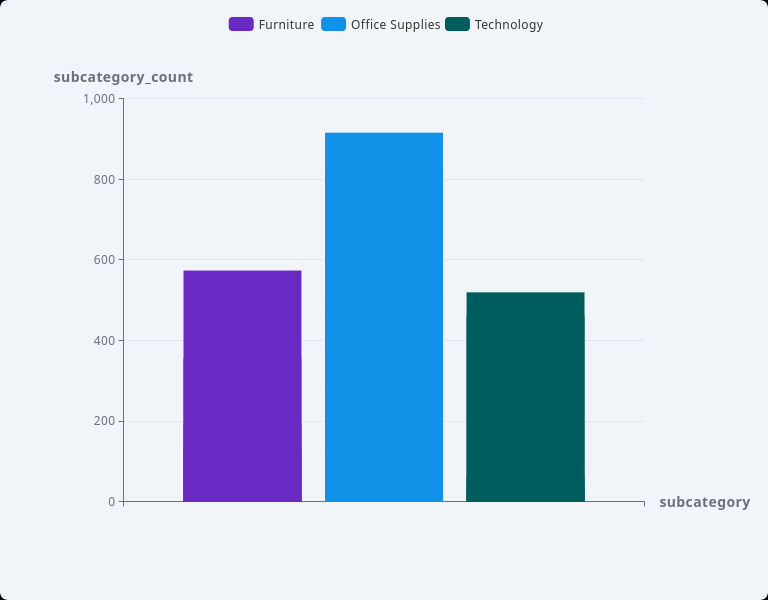}\\
 \hline
Show me a (radar chart) comparison of sales for each category per month & Radar Chart & \includegraphics[width=0.31\textwidth]{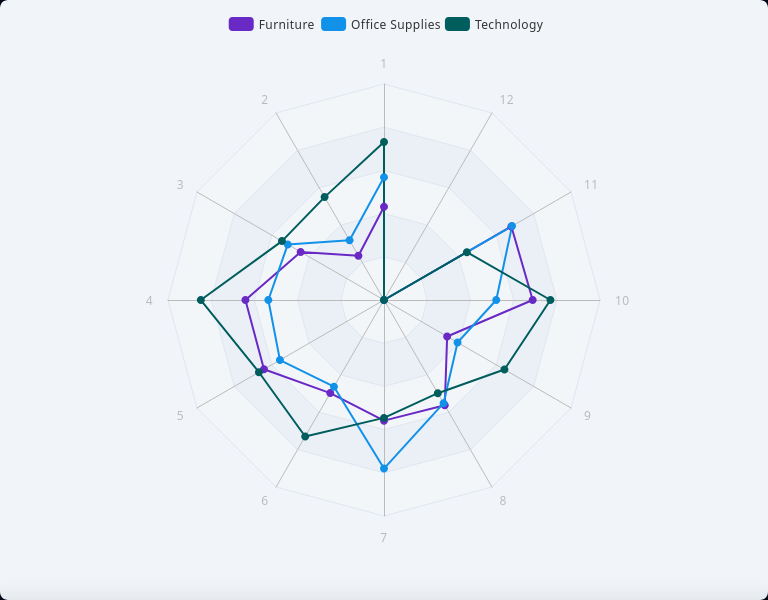}\\
 \hline
\end{tabularx}
\end{table}

\begin{table}
\caption{Generated charts using the LLM pipeline}
\label{table:results2}
\begin{tabularx}{1\textwidth} { 
  | >{\raggedright\arraybackslash}X 
  | >{\raggedright\arraybackslash}X 
  | >{\raggedright\arraybackslash}X | }
 \hline
\begin{center} \bf Prompt \end{center} & \begin{center} \bf Chart Name \end{center} & \begin{center} \bf Visualization \end{center} \\
\hline
Show me a comparison of sales for each category between (inclusive) 12/1/17 and 12/30/17, include the dates  & Line Chart & \includegraphics[width=0.31\textwidth]{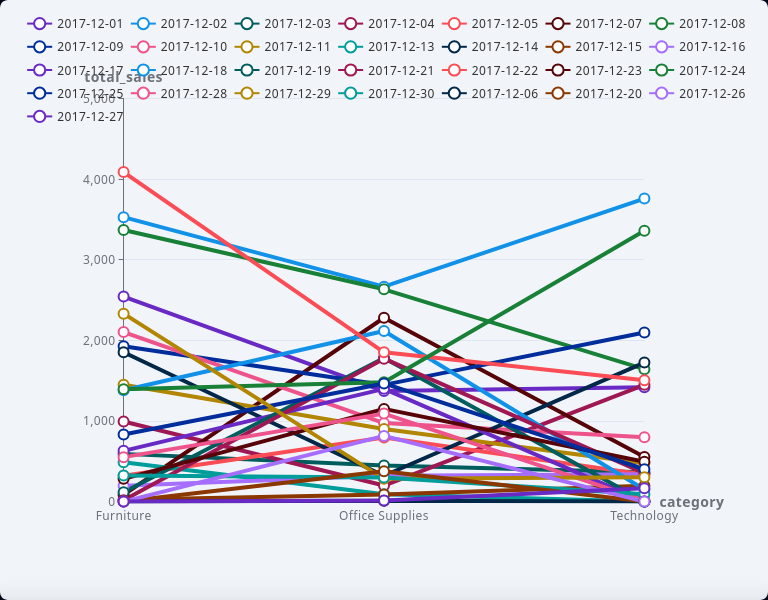}\\
 \hline
Show me a comparison of sales for each category between (inclusive) 12/1/17 and 12/3/17, include the date & Column Chart (2nd Method) & \includegraphics[width=0.31\textwidth]{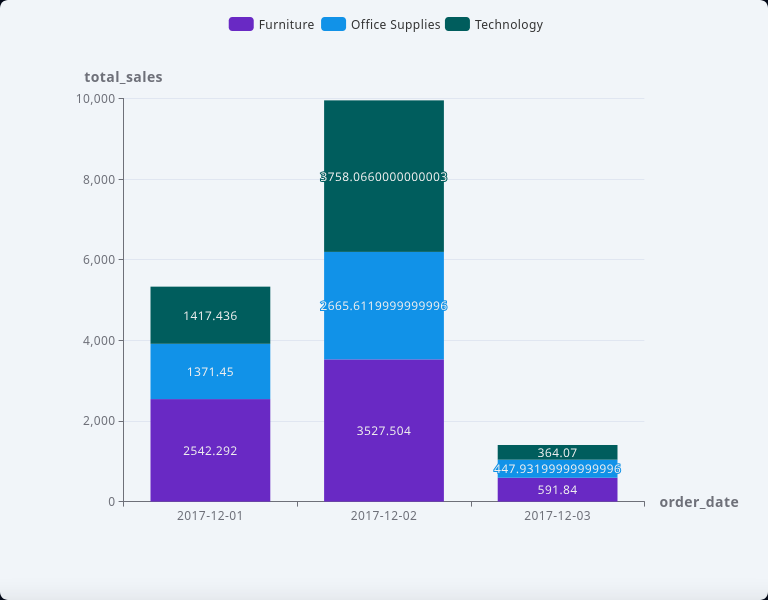}\\
 \hline
Show me a (line chart) comparison of sales for each city between (inclusive) 11/1/17 and 11/3/17, include the dates & Line Chart (2nd Method) & \includegraphics[width=0.31\textwidth]{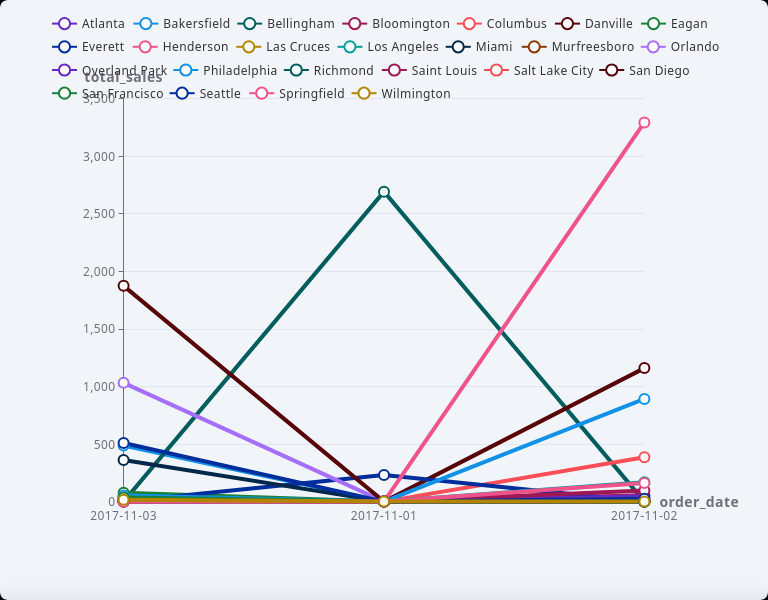}\\
\hline
Show me sales and profits & Scatter Chart & \includegraphics[width=0.31\textwidth]{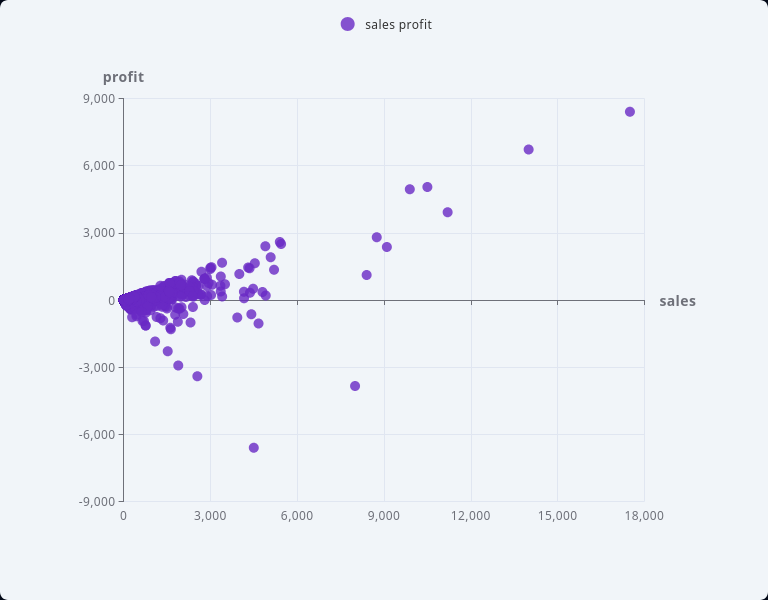}\\
 \hline
\end{tabularx}
\end{table}

\begin{table}
\caption{Generated charts using the LLM pipeline}
\label{table:results3}
\begin{tabularx}{1\textwidth} { 
  | >{\raggedright\arraybackslash}X 
  | >{\raggedright\arraybackslash}X 
  | >{\raggedright\arraybackslash}X | }
\hline
Show me the sales, sales forecast, and profit & Bubble Chart & \includegraphics[width=0.31\textwidth]{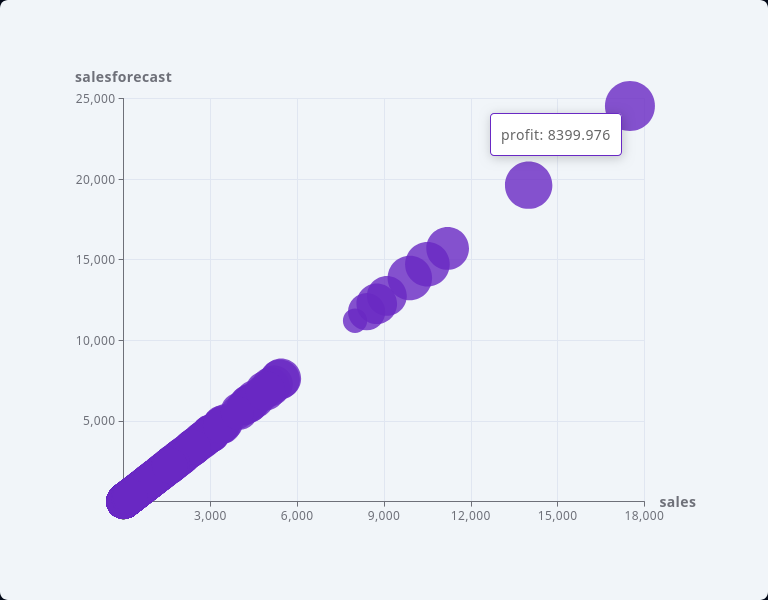}\\
 \hline
Show me the all quantities & Column Histogram & \includegraphics[width=0.31\textwidth]{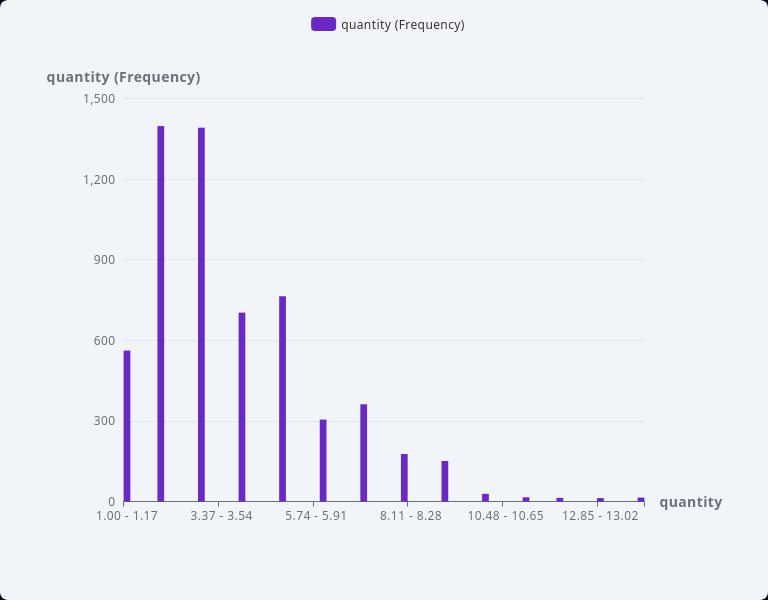}\\
 \hline
Show me all sales forecast & Line Histogram & \includegraphics[width=0.31\textwidth]{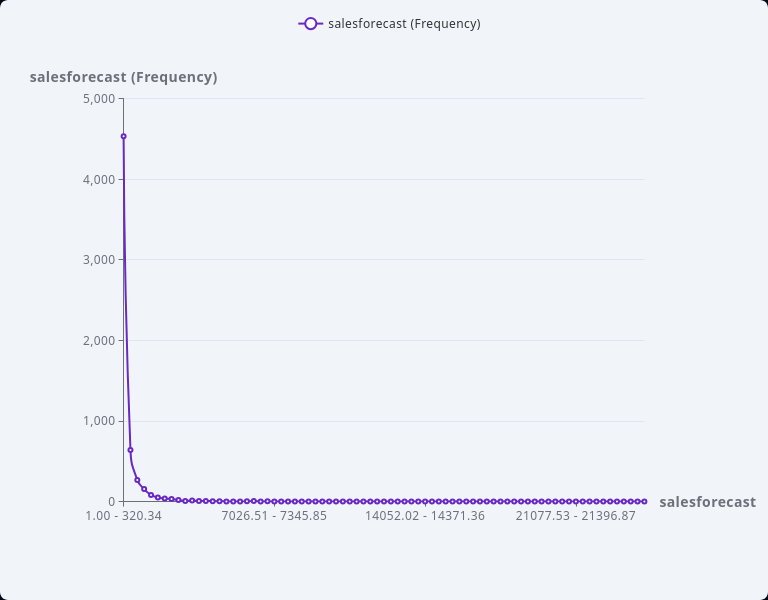}\\
 \hline
Show me the distribution of profit and days to ship, do not use analytical functions or sql suggestions & Scatter Chart (2nd Method) & \includegraphics[width=0.31\textwidth]{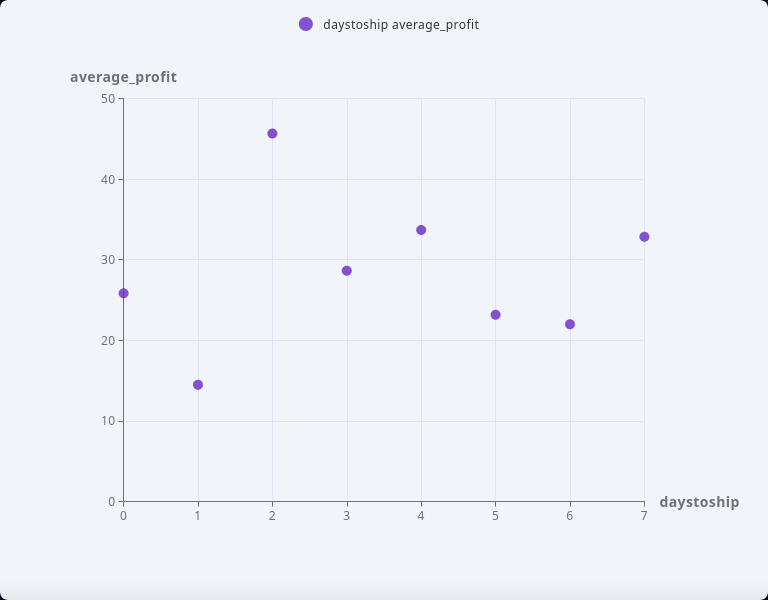}\\
 \hline
Show me the distribution between sales, sales forcast, and profit, do not use sql suggestions & 3D Area Chart & \includegraphics[width=0.31\textwidth]{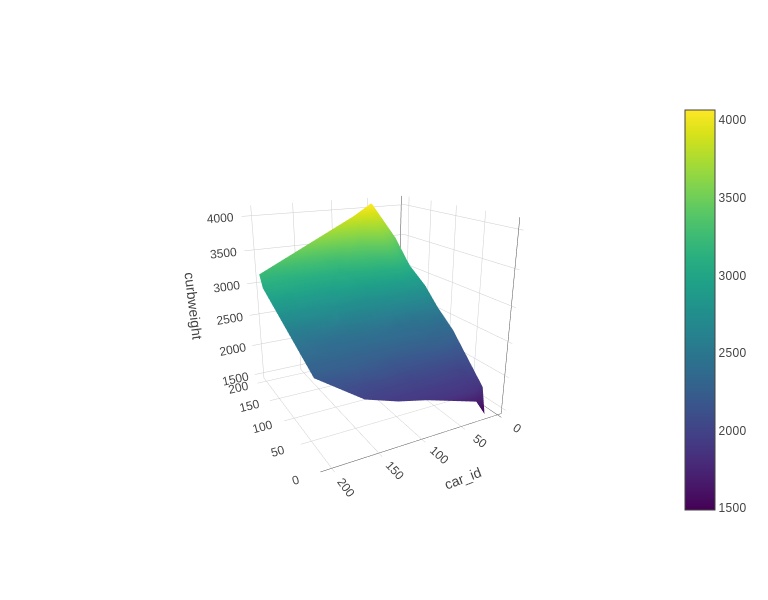}\\
 \hline
\end{tabularx}
\end{table}

\begin{table}
\caption{Generated charts using the LLM pipeline}
\label{table:results4}
\begin{tabularx}{1\textwidth} { 
  | >{\raggedright\arraybackslash}X 
  | >{\raggedright\arraybackslash}X 
  | >{\raggedright\arraybackslash}X | }
 \hline
 Show me the composition of percent in sales for each ship status for the dates 12/1/17 and 12/5/17, include the dates & Stacked 100\% Column Chart & \includegraphics[width=0.31\textwidth]{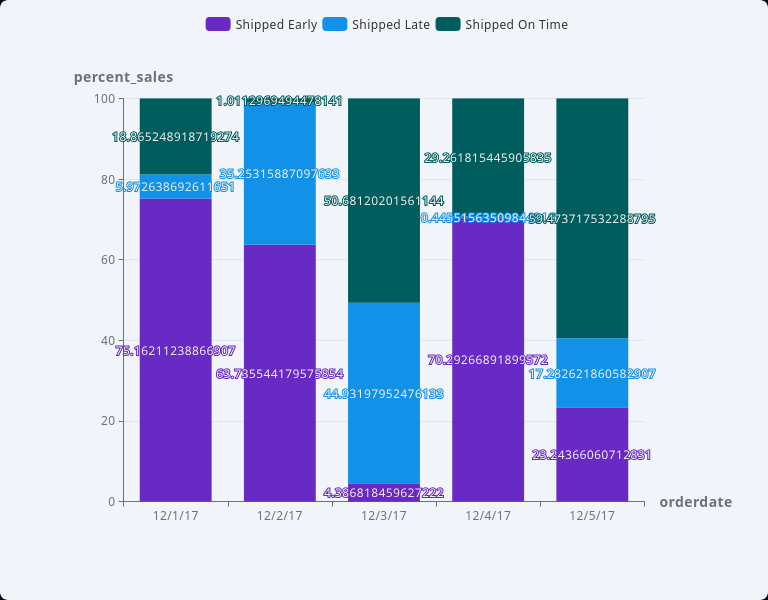}\\
  \hline
Show me the composition sales for each ship status for the dates 12/1/17 and 12/5/17, include the dates & Stacked Column Chart & \includegraphics[width=0.31\textwidth]{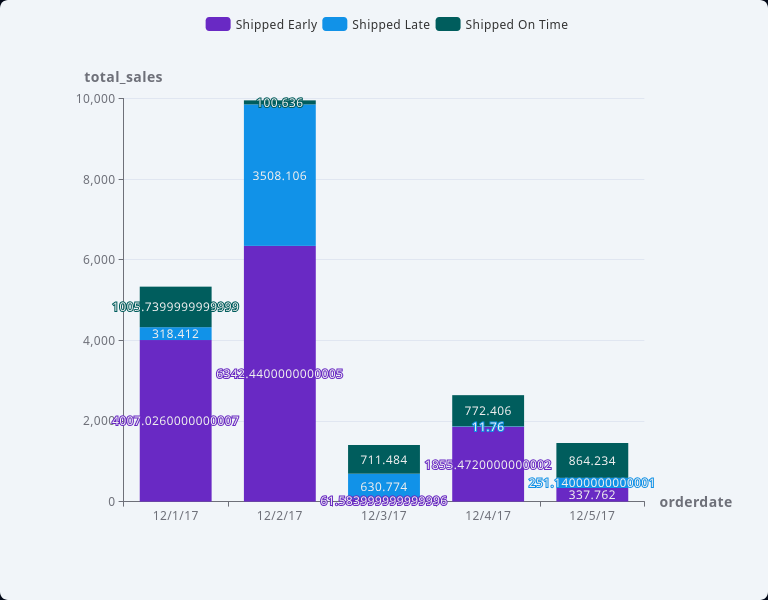}\\
 \hline
Show me the composition of sales for each category per month & Stacked 100\% Area Chart & \includegraphics[width=0.31\textwidth]{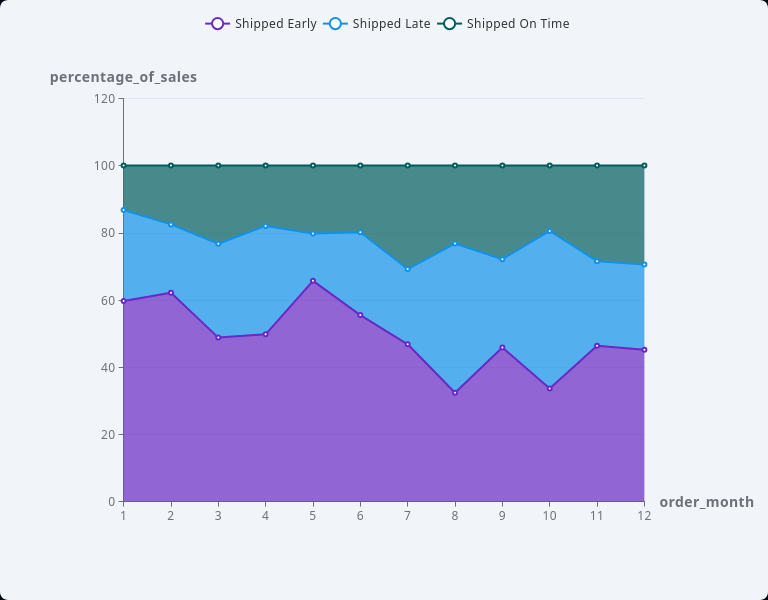}\\
 \hline
Show me the composition of percentage of sales for each ship staus per month & Stacked Area Chart & \includegraphics[width=0.31\textwidth]{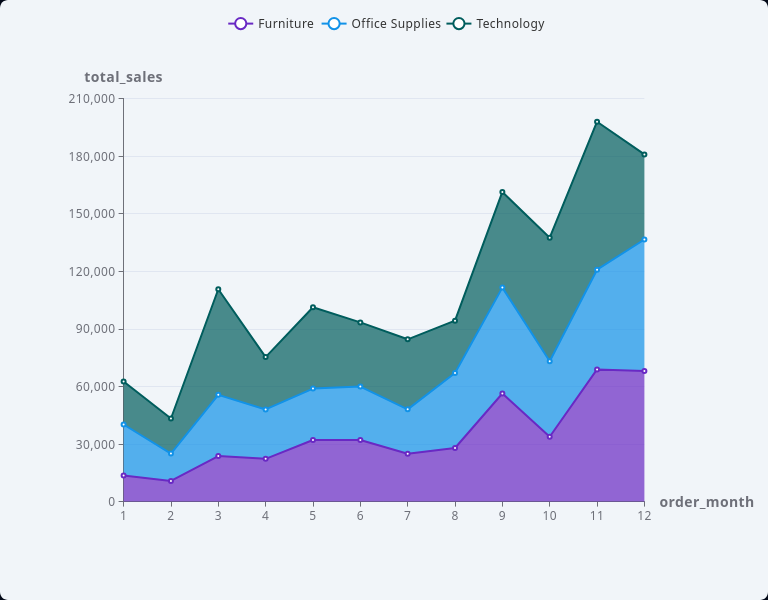}\\
 \hline
\end{tabularx}
\end{table}

\begin{table}
\caption{Generated charts using the LLM pipeline}
\label{table:results5}
\begin{tabularx}{1\textwidth} { 
  | >{\raggedright\arraybackslash}X 
  | >{\raggedright\arraybackslash}X 
  | >{\raggedright\arraybackslash}X | }
 \hline
 Show me the composition sales for each category & Pie Chart & \includegraphics[width=0.31\textwidth]{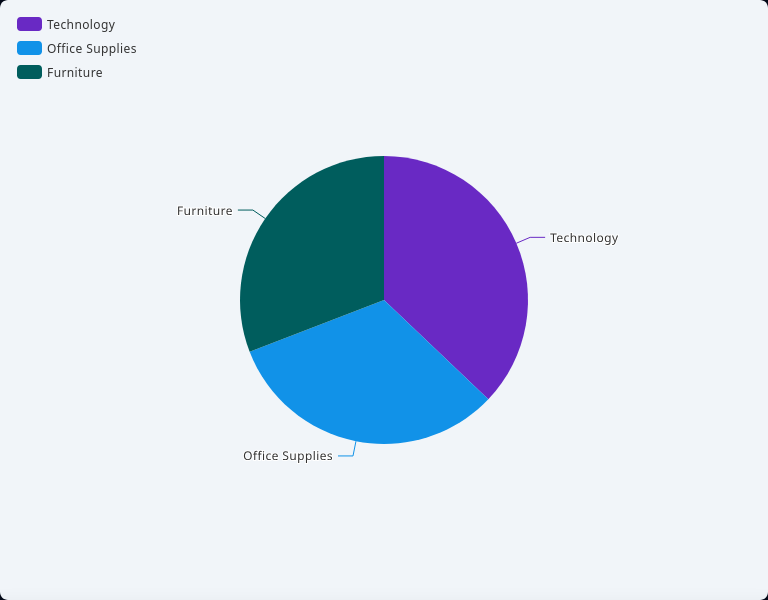}\\
 \hline
Show me the (waterfall chart) change in profit for ship status & Waterfall Chart & \includegraphics[width=0.31\textwidth]{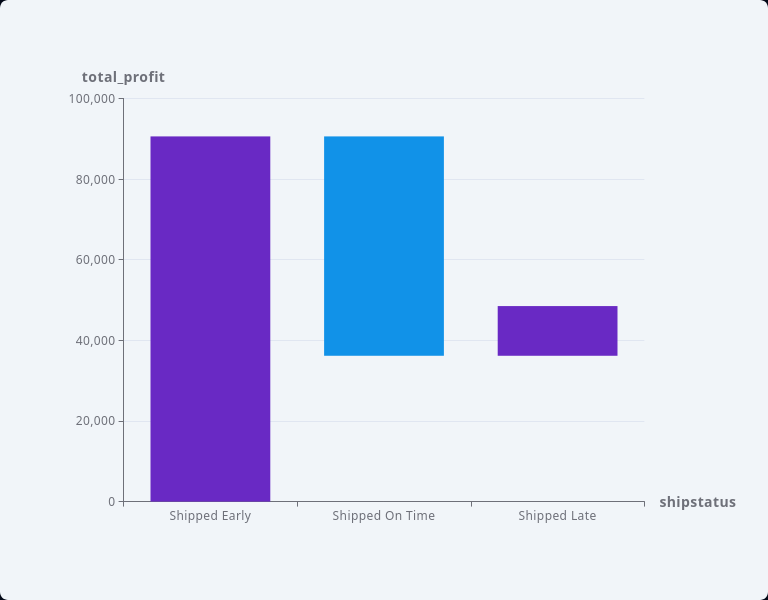}\\
 \hline
Show me the hierarchy of sales forecast for cities in the country (include country) & Treemap Chart & \includegraphics[width=0.31\textwidth]{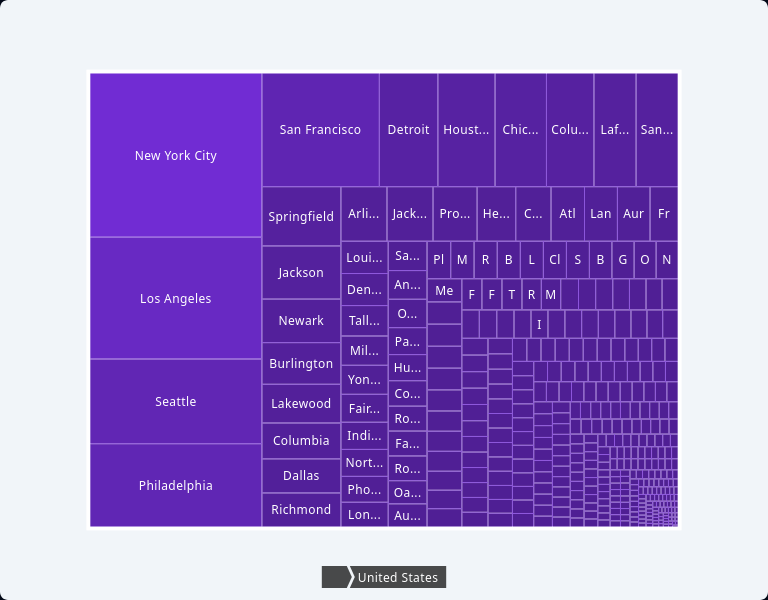}\\
 \hline
\end{tabularx}
\end{table}

% \paragraph{Using Abela's Taxonomy} When utilizing the fine-tuned LLM to select charts, we only witnessed 11 of the 21 charts being selected by the LLM.  These included: Treemap, Stacked Column Chart, Bar Chart, Stacked Area Chart, Line Histogram, Variable Width Column Chart, Waterfall Chart, Bubble Chart, Column Chart, Column Histogram, Scatter Chart.  Alternatively, if the user bypasses the LLM chart selector step, it will return a list of all possible charts.  This is by no means  an exhaustive test, and the prompts used for this specific test were generated with an LLM.

% \begin{table}
% \caption{Prompts used in testing}
% \label{table:macroquery}
% \begin{tabularx}{1\textwidth} { 
%   | >{\raggedright\arraybackslash}X 
%   | >{\raggedright\arraybackslash}X 
%   | >{\raggedright\arraybackslash}X | }
%  \hline
%  example & is macro-query? & reason \\
%  \hline
%  "which is the most affordable car?" & yes & The prompt does not directly reference any variables and requires sorting.  \\
%  \hline
%  "show me the car name with the lowest price" & no & The input attribute price is mentioned, but \\
%  \hline
%  "show me the car name for the least affordable car" & yes & The attribute referenced is the output attribute and not input attribute (price) which would have been used for sorting.  \\
% \hline
% \end{tabularx}
% \end{table}

\label{sec:headings}
\subsection{Observations}
\paragraph{Data Augmentation:}In the case of missing data, such as country of origin for motor-vehicles; we noticed that the LLM typically injects additional data.  As in the case for "what are the most bang for the buck european cars?" in Table~\ref{table:macroresultscar} the LLM would typically select from a pool of vehicle brands associated with European automakers in the SQL query.  This shows the potential to perform queries slightly outside the scope of the dataset.

\paragraph{Derived Metrics:} In the case of "bang-for-buck" and "0 to 60" for Tables~\ref{table:macroresultscar} and \ref{table:macroresultscar2}, the LLM has shown capabilities in deriving substitute metrics for unknown or missing data.  Potentially there is room for improving this behaviour by introducing web based resources for relevant formulas may result in better derived metrics.

\paragraph{Visualizing Analysis:}During our internal discussions; we noticed that a chart or table response to certain macro-queries were inappropriate.  Take for instance, when inquiring about correlation, LLM pipeline produces a single value as the correlation coefficient of two groups from the PostgreSQL's correlation function.  This can result in a table displaying a single row which proves problematic if the user's desire was to view a scatter plot visualization.  In future iterations, this could be mitigated by having two separate transformations pathways running asynchronously.  One for visualization generation which will forego any SQL analytical functions and a transformation to produce a simple single value response.  Additionally, for linear regression, it would be benefits to provide a scatter-plot with the best fit line superimposed on the chart.

% Issue has been fixed, there cannot be a single entry chart anymore
% This can result in a chart displaying a single data point, which proves problematic and would be better if the LLM produced both a heat-map (not currently available with our system) and a response of the correlation value.

\paragraph{AI Knows Best:}During our development, we noticed a noteworthy behaviour created by our prompts.  LLMs consistently overwrites user choice.  Suppose the user asks for a specific bar chart, however the LLM believes that a scatter chart is more appropriate and will not adhere to the user's request.  Sometimes, it may be impossible given the transformed CSV file containing unequal attribute counts compared to the input CSV.  Other times, the AI believes it is correct in overwriting user preference.  The following presents the output of one such instances when seeking to compare horsepower, price and car name:

\texttt{"Given the user's request to compare horsepower, price, and car name, it is clear that including car names in a chart with continuous variables would not be practical due to the high number of unique values. Instead, focusing on the relationship between horsepower and price would provide a clearer and more interpretable visualization."}

% .  However, this causes our method to generate a chart displaying only one entry.  This is evident in the scenario where asking for correlation, our system produced a graph with a single value due to results PostgreSQL's correlation function call.  Similarly, this is issue is reflected when attempting linear regression.
% % while ChatGPT produced a Correlation Matrix Chart

% FIX ME
% Furthermore, we discussed the potential of obfuscating certain portions of the response based on if the user has provided a macro-query or a more direct query.  We assume the extensive detailed response may prove overbearing to those who seek a straight forward answer with simplified justification.

% Sometimes user will want a chart type but its not really the best chart type
% TALK ABOUT OCCASSIONALLY USER SUGGESTED CHART BY AI SAYS NO TO IT

% PROTOTYPE2 ALIGNing w/ andrew abela's?

% EXCEL OF LIST OF PROMPTS TESTED
% MAKE MENTION OF CAR DATASET!  The results aligned with the dataset, if certain fields did not exist, the LLM generates them (i.e., british cars/ geolocation)
% Im feeling sad, what movies that watch?

\section{Conclusion}

This paper aims to highlight the important distinction of macro-queries in the context of data exploration, especially for broad categories of stakeholders such as policymakers, decision-makers, and the general public. Since most users may not have the ability to articulate or utilize attributes in the data effectively, macro-queries will be a crucial concept across the sciences to enhance interdisciplinary collaborations.  Furthermore, we constructed a prototype LLM pipeline capable of handling macro-queries and generate a conditionally diverse set of charts with Abela's Taxonomy\cite{abela}.  However, we do acknowledge that more work is required to improve results.

% \subsection{Future Direction}
% We anticipate to amalgamate the prior fine-tuning approach, as aforementioned, into the chart selection module to invoke a higher frequency of non-standard charts being chosen.  Due to its decent performance, of []] accuracy, we remain confident that this solution will yield desirable results when applied to our Chart Selection module to improve the quality of charts selected.  Similarly, using a specialized, fine tuned, model to convert user requests into SQL queries may prove beneficial.

% We plan to procure a more robust validation by developing empirical tests for direct queries where facts are non-opinionated and numerically verifiable such as "most fuel efficient four door sedans." and relying on a user acceptance survey for responses for opinionated macro-queries where individuals typically assign varying degrees of significance to attributes, with some emphasising priority of certain factors over others.

% At the start of our development, we noticed that prompting ChatGPT with macro-queries resulted in a drawn out conversation required the user manually prompt ChatGPT to make corrections.  However, at the conclusion of our development, we noticed that ChatGPT has improved and reliably creates a limited collection of charts with no corrections required.

% results are good enough

\section*{Acknowledgments}
This work was supported in part by National Science Foundation awards 2149133 and 2004014.

%Bibliography
\bibliographystyle{unsrt}
\bibliography{references}

\end{document}